\documentclass[]{TEAI}
\usepackage{helvet}

\usepackage{amsmath} 
\usepackage{natbib}
\usepackage{graphicx}
\usepackage{subcaption} 

\usepackage[toc,page,header]{appendix}
\usepackage[utf8]{inputenc} 
\usepackage[T1]{fontenc}    
\usepackage{hyperref}       
\usepackage{url}            
\usepackage{booktabs}       
\usepackage{lmodern}        
\usepackage{amsfonts}       
\usepackage{nicefrac}       
\usepackage{microtype}      
\usepackage{wrapfig}

\usepackage{amssymb}  
\usepackage{fontawesome}  
\usepackage{url}  

\usepackage{titletoc}

\usepackage{tikz}  
\usepackage{comment}  
\usepackage{tabularx}  
\usepackage{booktabs}  

\usepackage{minitoc}

\usepackage{booktabs}
\usepackage{array}
\usepackage{etoolbox}

\definecolor{lightblue}{RGB}{200, 230, 255}  
\definecolor{headerblue}{RGB}{150, 200, 255} 

\usepackage{pgfplots}
\usepackage[utf8]{inputenc} 
\usepackage[T1]{fontenc}    
\usepackage{hyperref}       
\usepackage{url}            
\usepackage{booktabs}       
\usepackage{amsfonts}       
\usepackage{nicefrac}       
\usepackage{microtype}      
\usepackage{xcolor}         
\usepackage{graphicx}
\usepackage{float}
\usepackage{comment}
\usepackage{multirow} 
\usepackage{amsmath} 
\usepackage{makecell} 
\usepackage{siunitx}  
\usepackage{tikz}
\usepackage{pgf-pie} 
\usepackage{subcaption}
\usepackage{wrapfig}
\usepackage[export]{adjustbox}

\usepackage{ragged2e}      
\usepackage{tabularx}       
\usepackage{array}          
\usepackage{caption}        
\usepackage{enumitem}
\usepackage{pifont}
\usepackage[hang,flushmargin]{footmisc} 

\usepackage{tcolorbox}

\usepackage{tcolorbox}    
\tcbuselibrary{breakable}  
\tcbuselibrary{skins}      

\usepackage{tabularx}
\usepackage{listings}
\usepackage{algorithm}
\usepackage{algorithmic}

\title{\textsc{Real In, Real Out}: What If We Only Use Real Data for Scene Text Editing?}

\author{
    Xingsong Ye\textsuperscript{1,2},
    Yongkun Du\textsuperscript{1,2},
    Jiaxin Zhang\textsuperscript{3},
    Chong Sun\textsuperscript{3},
    Zhineng Chen\textsuperscript{1,2,$\dagger$}
}

\affiliation[1]{\mbox{Institute of Trustworthy Embodied AI, Fudan University}}
\affiliation[2]{\mbox{Shanghai Key Laboratory of Multimodal Embodied AI}}
\affiliation[3]{\mbox{WeChat Vision, Tencent Inc.}}

\abstract{
\begin{abstract}

Aiming to modify textual content while preserving the original style, Scene Text Editing (STE) has significant practical value in many applications. STE typically requires paired images before and after editing for training.  However, recent approaches rely heavily on synthetic paired data and complex disentanglement strategies. They struggle to faithfully capture the diversity and complexity of natural scenes, often producing artifacts or hallucinations.  In this work, we challenge the common practice and investigate whether STE can be driven by turning each real sample into its own paired supervision. Concretely, we explore multiple ways to disrupt the original image (e.g., Crop, Shuffle, Mask) to construct self-paired training data, and build a strong baseline (\textbf{TextRIRO}) upon a conditional diffusion architecture. Our experiments show that combining character-column shuffling with image cropping (Shuffle\&Crop) effectively destroys text semantics while preserving key style cues, enabling the model to learn editing behaviors directly from the real text image. 
With this concise yet real-oriented design, TextRIRO achieves promising performance in both accuracy and visual realism across real-world benchmarks, substantially mitigating the synthetic-to-real domain gap. Furthermore, we leverage TextRIRO to build \textbf{TextRIRO-3M}, a large-scale realistic dataset for long-text recognition by concatenating edited and original images. Experimental results show that training recognition models with TextRIRO-3M significantly improves recognition accuracy, demonstrating that TextRIRO provides highly useful training resources for downstream tasks. Overall, TextRIRO lifts STE to realistic, reliable, and widely applicable new levels.
\end{abstract}
}

\correspondence{\email{zhinchen@fudan.edu.cn}}
\checkdata[Code]{\url{https://github.com/YesianRohn/TextRIRO}}

\begin{document}
\maketitle
\renewcommand{\thefootnote}{}
\footnotetext{\\$^\dagger$Corresponding authors.}
\footnotetext{Accepted as a conference paper at NeurIPS 2026.}
\renewcommand{\thefootnote}{\arabic{footnote}}

\vspace{-1.5em}

\section{Introduction}
\label{sec:intro}

\begin{figure}[t]
 \centering
 \includegraphics[width=0.95\linewidth]{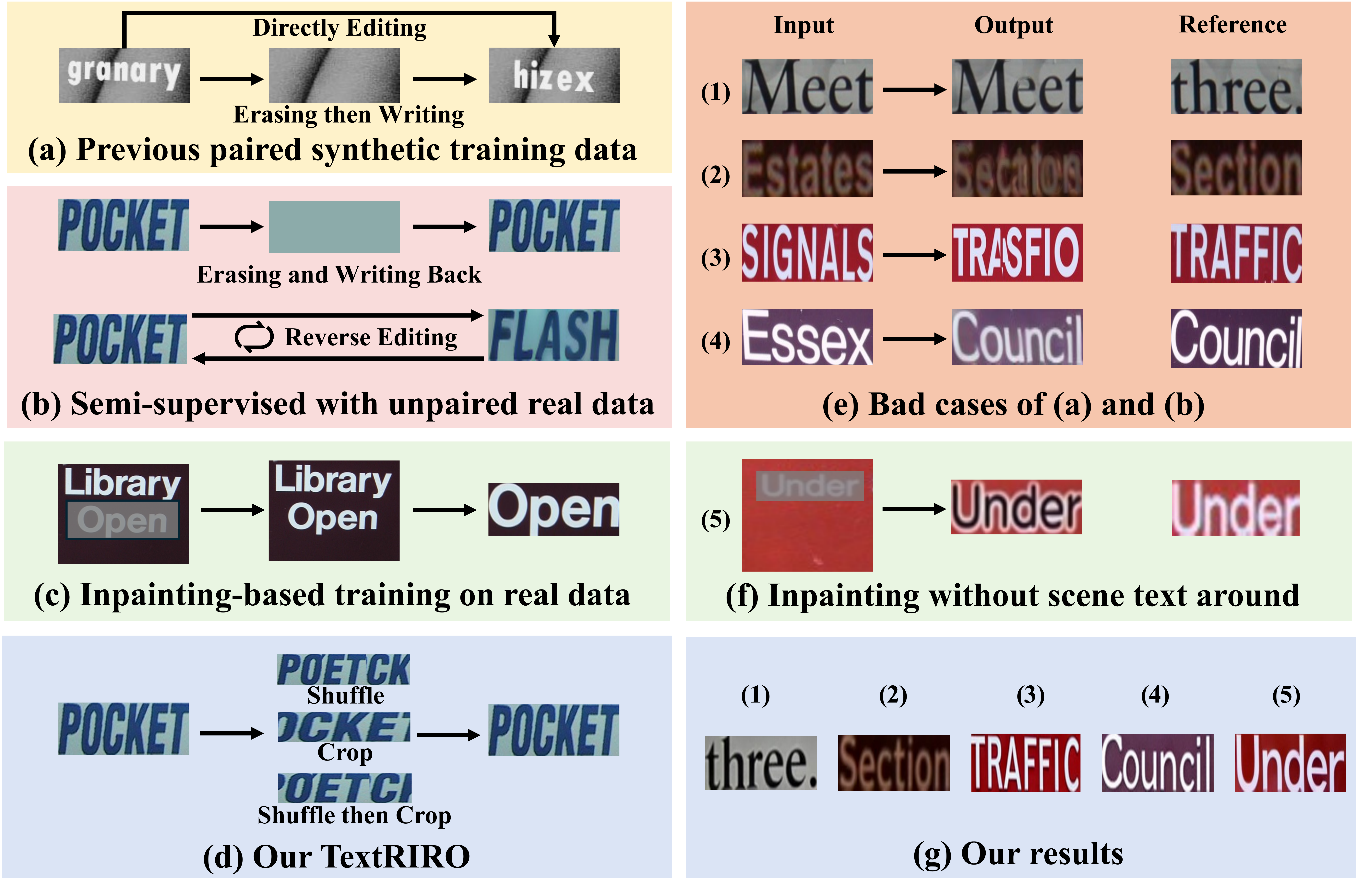}
 \caption{\textbf{A data-centric view of Scene Text Editing.} The comparison of the previous methods and TextRIRO between the data utilization paradigm (left) and the editing results (right). In (e), we present the artifacts and hallucinations of the wrong editing results in (a) and (b).}
 \label{fig:overview}
\end{figure}

Scene Text Editing (STE) aims to extract the style of the given text and replace it with a new string, while maintaining both visual realism and style consistency~\cite{wu2019editing}.
STE is typically formulated as a generative task, and its applications grow with the advancement of generative models. 
It addresses practical needs such as replacing text in images for artistic creation and also facilitates various downstream scene text tasks. 
Notably, STE shows significant potential in generating data~\cite{textssr} for scene text recognition (STR)~\cite{duijcai2022svtr, zheng2024cdistnet, ccd_guan, ccdplus_guan}, as it can generate samples with the same style but different content from existing images.

However, it is hardly possible to get different texts with the same background style in real-world scenarios. To address this, previous methods~\cite{wu2019editing, yang2020swaptext, roy2020stefann, krishnan2023textstylebrush, zeng2024textctrl} rely heavily on synthetic data, as shown in Fig.~\ref{fig:overview}(a). On the one hand, two-stage explicit disentanglement methods~\cite{wu2019editing, yang2020swaptext, roy2020stefann} use rendering tools to synthesize paired images with identical style but different text over a background image, followed by foreground-background separation and fusion. But they fail to separate text from the background, resulting in incomplete removal of the source text, as shown in cases (1) and (2) in Fig.~\ref{fig:overview}(e). On the other hand, implicit disentanglement approaches~\cite{krishnan2023textstylebrush, zeng2024textctrl} leverage various pre-trained models to extract features for directly generating target images, which is less controllable, resulting in hallucinations such as case (3), with misspelled or missing characters. Recent efforts~\cite{mostel, lee2021rewritenet, rsste} have explored semi-supervised learning with real data by erasing itself and writing back or paired training using pseudo-labeled data, as illustrated in Fig.~\ref{fig:overview}(b). Nevertheless, synthetic data still remains dominant. As evident from the editing results in Fig.~\ref{fig:overview}(e), this branch of methods easily leads to case (4): even when the target text is generated, it often appears mismatched in style or contains visible artifacts. Furthermore, inpainting-based STE paradigms (Fig.~\ref{fig:overview}(c)) mask the text region and attempt to restore it using surrounding context~\cite{diffste, chen2023diffute, textssr, Chen_2026_CVPR}. However, the lack of sufficient style cues makes it difficult to recover the desired appearance, as shown in Fig.~\ref{fig:overview}(f).

We are motivated by an observation that prior work has under-emphasized: \emph{the domain gap in STE is primarily a data problem, and only secondarily an architecture problem}.  We therefore ask the simplest possible version of the question:
\begin{quote}
  \textit{What if we only use real data for scene text editing, with neither synthetic triplets nor expensive paired collection?}
\end{quote}

The core challenge is to extract stable style guidance from the source text image while suppressing the side effect from semantic text content.  Our key insight is that scene text style mainly lies in low-level cues such as color, texture, and illumination, while semantic text is a high-level sequential structure.  Based on this, we explore multiple self-paired real-data schemes that force the model to focus on style rather than semantics. Specifically, we experiment with (i) shuffling and recombining the text region either column-wise or block-wise (Shuffle), (ii) cropping local sub-regions (Crop), and (iii) masking parts of the original image (Mask). Empirically, we find that combining column-wise Shuffle with Crop (Shuffle\&Crop) achieves the best performance.

To realize the idea, we propose \textbf{TextRIRO}, a conditional diffusion model that takes only \textbf{R}eal data as \textbf{I}nput and produces \textbf{R}ealistic \textbf{O}utput for scene \textbf{Text} editing. During training, TextRIRO applies ``Shuffle\&Crop'' to a real text image to separate its style and semantics, and then reconstructs the original image. By removing the reliance on synthetic data, it overcomes the bottleneck of high-quality generation during inference. For style features, we process them into multi-scale representations and then inject them into the conditional U-Net using the ControlNet-like process~\cite{controlnet} for efficient integration. For text encoding, we use an STR pre-trained encoder~\cite{jiang2023revisiting} to extract character-level features, which improves text understanding and enables robust STE transfer.

TextRIRO removes the need for explicit or implicit disentanglement and avoids using the synthetic domain as the main training support. On real benchmarks for STE, it produces more realistic results with markedly reduced ``synthetic flavor.'' It also achieves higher accuracy on equal-length editing and variable-length editing tasks. These results show that self-paired training on real data effectively moves STE toward realistic generation. We further ask whether STE can help STR data generation~\cite{textssr,ye2026wrong,ye2026advancing}. We find that TextRIRO already outperforms existing STE methods, but the editing accuracy still limits direct use as training data. We then use STE as data expansion by concatenating the original image with its edited counterpart. This preserves style but adds content, which yields high-quality, scene-consistent samples for long-text recognition. Starting from the labeled real dataset U14M-Filter~\cite{jiang2023revisiting, du2024svtrv2}, we build a 3M-scale dataset, \textbf{TextRIRO-3M}. On Union14M-Benchmark~\cite{jiang2023revisiting}, training an STR model on mixed TextRIRO-3M and real data improves overall accuracy by 1.91\% over a real-only baseline, and improves the Multi-words subset by 3.28\%. These gains indicate that current STR training data are not saturated and STE-based synthetic data substantially boost STR.

In summary, our main contributions are as follows:
\begin{itemize}
 \item We propose TextRIRO, a novel STE method that trains solely on real data via self-paired supervision. It performs realistic editing without explicit or implicit disentanglement, and it alleviates artifacts and hallucinations caused by the synthetic-to-real domain gap.
 \item TextRIRO achieves more realistic and accurate edits on real benchmarks and attains state-of-the-art (SOTA) performance under the stronger criterion of ``training with edited data.''
 \item Built on TextRIRO, we create TextRIRO-3M, a large-scale realistic composite dataset for long-text recognition. It significantly improves STR, which validates both the realism and the utility of our edited data and demonstrates that STE effectively benefits STR.
\end{itemize}

\section{Related Work}

\paragraph{Model Perspective on STE}

Traditional STE methods evolve from GAN-based models~\cite{wu2019editing, yang2020swaptext, roy2020stefann, krishnan2023textstylebrush} to diffusion models~\cite{diffste, chen2023diffute, anytext, udifftext, tuo2024anytext2, zeng2024textctrl} and, more recently, to Unified Multimodal Models (UMMs)~\cite{zhang2024choose, rsste}. Early STE models employ GANs to directly edit foreground text, as seen in SRNet~\cite{wu2019editing} and its multi-stage framework for reconstructing background, rendering text, and fusing outputs. Later works expand capabilities: SwapText~\cite{yang2020swaptext} handles curved text, STEFANN~\cite{roy2020stefann} focuses on character-level control, and TextStyleBrush~\cite{krishnan2023textstylebrush} leverages StyleGAN2~\cite{karras2020analyzing} for self-supervised style-content disentanglement. Diffusion models advance STE further: DiffSTE~\cite{diffste} uses dual encoders, DiffUTE~\cite{chen2023diffute} introduces inpainting and OCR-guided encoding, TextDiffuser~\cite{chen2023textdiffuser} and UDiffText~\cite{udifftext} use segmentation masks for better generation. Frameworks like AnyText~\cite{anytext} and AnyText2~\cite{tuo2024anytext2} unify multilingual editing via ControlNet~\cite{controlnet}. TextCtrl~\cite{zeng2024textctrl} disentangles style conditions directly in diffusion models. Meanwhile, DARLING~\cite{zhang2024choose} unifies scene text editing, removal, and recognition tasks to leverage these tasks to disentangle task-specific features and RS-STE~\cite{rsste} utilizes scene text recognition to assist scene text editing. These two approaches highlight synergy across scene text tasks.

\paragraph{Data Perspective on STE}

Model advancements in STE are closely tied to the evolution of training data. Early methods use synthetic paired data by rendering tools, e.g., SRNet-Datagen~\cite{wu2019editing}. To include real-world data, RewriteNet~\cite{lee2021rewritenet} uses masked regions and MOSTEL~\cite{mostel} proposes an erasing and writing-back strategy. UMM-based RS-STE~\cite{rsste} adopts reverse self-supervision on real images. In diffusion-based STE, some models (e.g., TextCtrl~\cite{zeng2024textctrl}) disentangle features using synthetic data, while others~\cite{anytext, chen2023textdiffuser} view STE as masked image inpainting, releasing large real datasets like Anyword-3M~\cite{anytext} and MARIO-10M~\cite{chen2023textdiffuser}. However, style consistency can be difficult when context lacks similar references. Benchmarks and datasets for application are also advancing: ScenePair~\cite{zeng2024textctrl} provides high-quality real-world evaluation, and TextSSR~\cite{textssr} generates millions of edited images for STR tasks, demonstrating STE's broader utility.
\begin{figure}[t]
 \centering
 \includegraphics[width=0.95\linewidth]{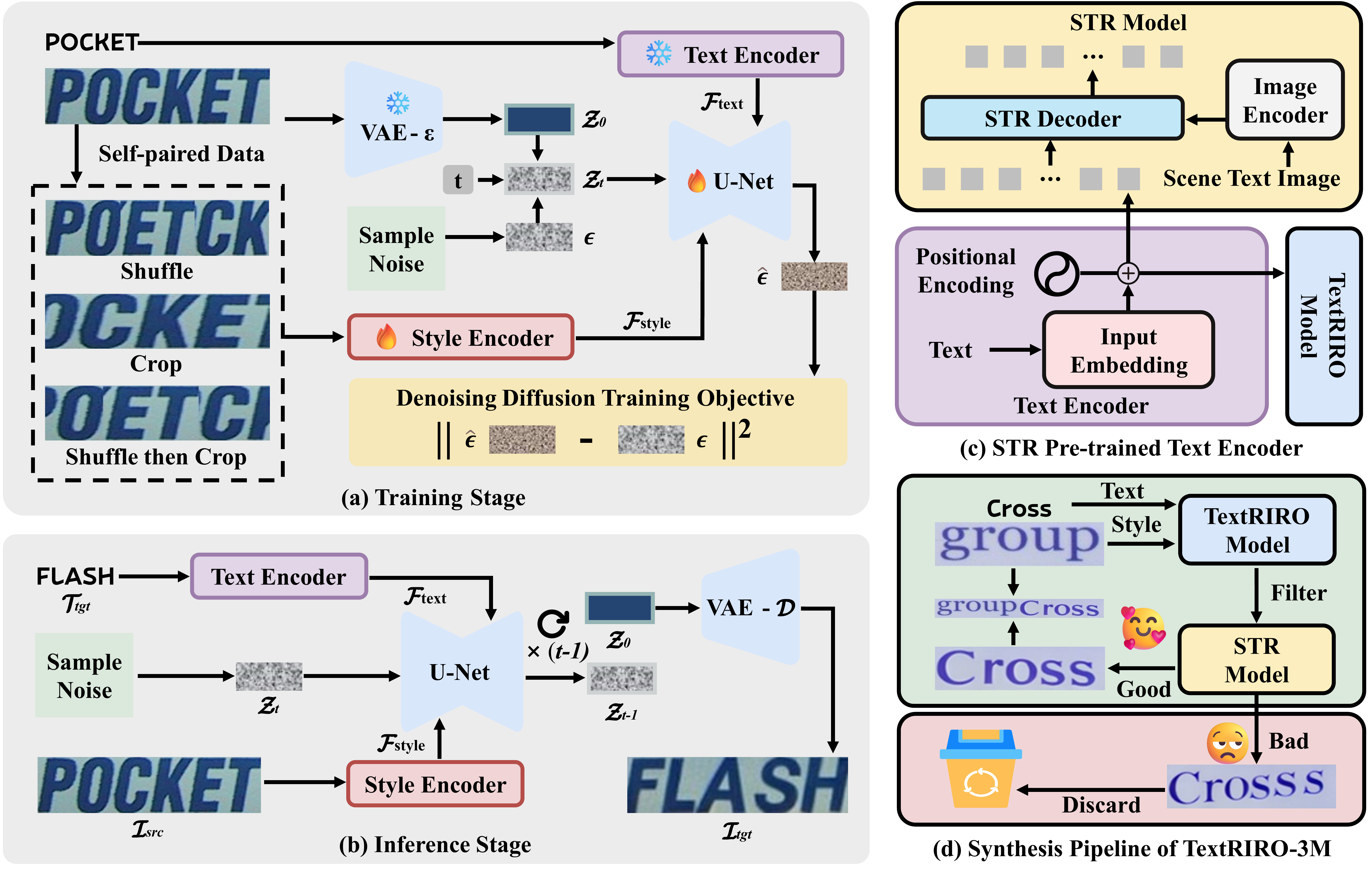}
  \caption{\textbf{TextRIRO pipeline.} (a) At training time, a single real image $I_{\text{tgt}}$ is pushed through a content-destructive, style-preserving operator to produce a pseudo-source $I_{\text{src}}$. A ViT-based style encoder emits multi-scale features that are injected into the denoising U-Net through ControlNet-style zero-convolutions, while a text encoder pre-trained on \emph{real} scene text with an autoregressive STR objective supplies the target-text conditioning.  (b) At inference time, the same U-Net follows the standard SD denoising schedule.  (c) Our text encoder is a lightweight character tokenizer trained jointly with an STR decoder via next-token prediction.  (d) The edited image is validated by an independent STR model and, if accepted, concatenated with its original to build a style-consistent long-text sample for STR.  \raisebox{-0.5em}{\includegraphics[height=1.3em]{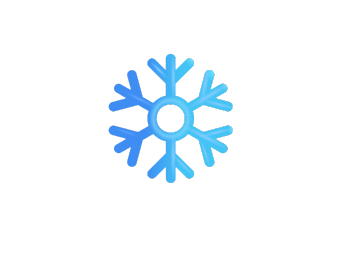}}: frozen, \raisebox{-0.5em}{\includegraphics[height=1.3em]{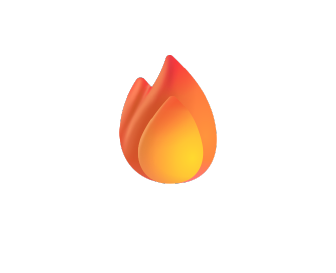}}: trainable.}
 \label{fig:model}
\end{figure}

\section{Method}

The STE process can be formally defined as follows. Denote the input source image as $I_{\text{src}}$, the target text as $T_{\text{tgt}}$, and the editing model as $M$. The objective is to produce the target image $I_{\text{tgt}}$ such that:
\begin{equation}
I_{\text{tgt}} = M(I_{\text{src}}, T_{\text{tgt}}).
\label{equation_aim}
\end{equation}
However, Eq.~\ref{equation_aim} characterizes the ideal training and inference procedure. In practice, training data rarely provide real triplets $(I_{\text{src}}, T_{\text{tgt}}, I_{\text{tgt}})$. To bridge this gap, Sec.~\ref{sec_3-1} introduces a self-paired training strategy, Sec.~\ref{sec_3-2} presents the unified TextRIRO architecture, and Sec.~\ref{sec_3-4} leverages the trained model to expand STR datasets with long-text instances.

\subsection{Self-Paired STE Data}
\label{sec_3-1}

Most prior data pipelines~\cite{wu2019editing, mostel} construct $(I_{\text{src}}, T_{\text{tgt}}, I_{\text{tgt}})$ triplets primarily from synthetic text images. The domain gap between synthetic and real scenes limits the ability of STE models and their applications to downstream scene text tasks. Recent methods~\cite{mostel, rsste} introduce some real data for unpaired training, yet synthetic samples remain dominant and only partially bridge this gap.

\begin{algorithm}[ht]
\caption{Self-Paired STE Data Generation}
\begin{algorithmic}[1]
\REQUIRE Scene text image $I_\text{tgt}$, corresponding text label $T_\text{tgt}$, shuffle probability $p_\text{shuffle}$
\ENSURE Generated self-paired source image $I_\text{src}$
\STATE Let $L = |T_\text{tgt}|$ denote the length of the text label.
\STATE Partition $I_\text{tgt}$ along its longest axis into $L$ equally sized segments: $\mathcal{B} = \{B_1, B_2, ..., B_L\}$.
\STATE Sample $r \sim \mathcal{U}(0, 1)$.
\IF{$r < p_\text{shuffle}$}
 \STATE Randomly permute the order of the segments to obtain $\mathcal{B}' = \{B'_1, ..., B'_L\}$.
\ELSE
 \STATE Set $\mathcal{B}' = \mathcal{B}$ (i.e., maintain original order).
\ENDIF
\STATE Choose $N \leq L$ and randomly select a consecutive subset of $N$ segments from $\mathcal{B}'$.
\STATE Concatenate the selected segments to form $I_\text{src}$.
\STATE \textbf{Return} $(I_{\text{src}}, T_{\text{tgt}}, I_{\text{tgt}})$
\end{algorithmic}
\label{alg_data}
\end{algorithm}

When using an image as $I_{\text{src}}$, our goal is to remove its textual content while preserving its style. This motivates a simple strategy: scrambling the original image disrupts content while retaining stylistic features. However, since text comprises discrete characters, excessive shuffling fragments glyphs and impairs shape consistency. This presents a trade-off: we aim to maximize content removal but minimize stylistic degradation. (1) Text typically aligns along its long axis, and its content and length are known. We exploit this structural prior by dividing the image along its long axis into $L$ ($L$ is the number of characters) equal bins. By reordering these bins, we generate the source image $I_{\text{src}}$ (see ``Shuffle'' in Fig.~\ref{fig:model}), thereby breaking semantic continuity while maintaining local appearance cues. (2) Stripe-based shuffling alone is insufficient for complex text layouts, such as curved or multi-oriented arrangements. To address this, we propose an alternative that reconstructs the image through its sub-patches. The global text arrangement is locally preserved: tilts and curvature manifest within contiguous sub-images. We form $I_{\text{src}}$ by cropping the original image (as illustrated by ``Crop'' in Fig.~\ref{fig:model}), yielding self-paired data that retain the stylistic integrity of intricate layouts. (3) Combining these two strategies results in a more robust and diverse self-paired mechanism. As shown in Fig.~\ref{fig:model} (``Shuffle then Crop''), with probability $p_\text{shuffle}$, shuffling is applied before cropping. Otherwise, cropping is applied directly to the original image. Alg.~\ref{alg_data} presents the complete algorithm of our self-paired dataset. The subsequent ablation study presents other self-paired methods (e.g., Crop Only, Shuffle Only, Mask) and results, further demonstrating the optimality of the adopted algorithm.

\subsection{Model Architecture}
\label{sec_3-2}
With the constructed self-paired triplets, TextRIRO maps the perturbed source image $I_{\text{src}}$ and target text $T_{\text{tgt}}$ into style and text control signals for a ControlNet-like diffusion generator~\cite{controlnet}, as shown in Fig.~\ref{fig:model}. The style encoder follows TextCtrl~\cite{zeng2024textctrl}: a ViT backbone~\cite{dosovitskiy2020vit} extracts image representations, and a lightweight spatial pyramid aggregates them into multi-scale features that are additively injected into the U-Net. For text, we keep character-level encoding~\cite{udifftext, zeng2024textctrl} but replace synthetic glyph/CLIP pre-training with an autoregressive STR objective on real scene text~\cite{sheng2019nrtr, jiang2023revisiting}. This improves real-domain character awareness and aligns recognition with editing.

The generator contains a VAE for latent mapping, a U-Net for denoising, and the above style/text encoders. The VAE and text encoder are frozen, while the style encoder and U-Net are jointly trained with the standard noise-prediction diffusion objective conditioned on style and text. This design is intentionally simple and backbone-agnostic: stronger diffusion backbones or conditioning modules can be substituted when more compute is available. We also evaluate a DiT-based conditional control alternative in Tab.~\ref{tab:control_backbone}, where ControlNet-style multi-scale injection remains more effective for fine-grained scene text editing. Detailed feature equations, tokenization, and the training loss are provided in Appendix Sec.~\ref{app:model_formulation}.

\subsection{Generating Long-Text Recognition Data}
\label{sec_3-4}
After training TextRIRO, we generate a style-matched image with the same length but different content for each real input. To ensure correctness, we filter generated samples with an STR model trained on real data and keep only instances whose predicted text matches $T_{\text{tgt}}$. We then concatenate each filtered image with its original reference to build long-text recognition data, doubling the sequence length. The detailed pipeline and visualizations are illustrated in Fig.~\ref{fig:model} (d). This procedure substantially broadens coverage for rare scenarios, such as long and multi-word instances, and it establishes a practical connection between STE and STR.

\begin{figure*}[t]
\centering
\includegraphics[width=\textwidth]{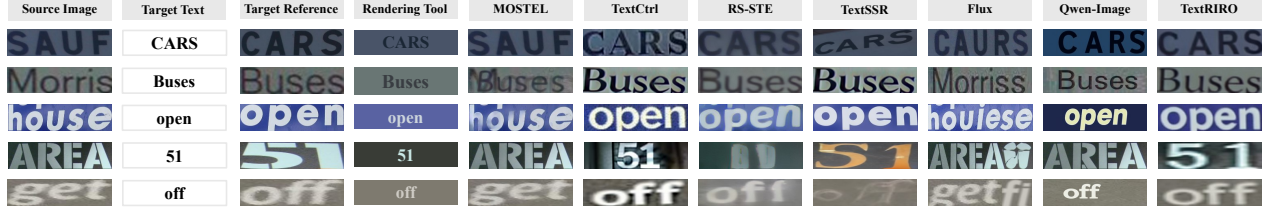} 
  \caption{\textbf{Qualitative comparison on ScenePair.}  ``Target Reference'' is the real paired ground truth. ``Rendering Tool'' composites the target string on the segmented foreground/background of the source and serves as an accuracy-only reference.  TextRIRO is the only method that consistently reproduces the stroke thickness, perspective and illumination of the reference without leaving a synthetic residue.}
\label{fig:comp_vis}
\end{figure*}
\section{Experiments}

\subsection{Editing Performance Comparison}
\noindent\textbf{Results on ScenePair Benchmark.}
We evaluate the proposed TextRIRO on the ScenePair~\cite{zeng2024textctrl} against both specialized STE methods and general editing models. The specialized baselines include the GAN-based SRNet~\cite{wu2019editing} and MOSTEL~\cite{mostel}, the diffusion-based inpainting methods AnyText~\cite{anytext} and TextSSR~\cite{textssr}, the direct diffusion editing method TextCtrl~\cite{zeng2024textctrl}, and the UMM-based RS-STE~\cite{rsste}. The general models include Flux.1 Kontext~\cite{flux} and Qwen-Image-Edit~\cite{wu2025qwen}. For these two models, we use a fixed prompt: ``Edit the text to `$T_{\text{tgt}}$','' where $T_{\text{tgt}}$ is the designated target. We follow the ScenePair protocol for evaluation. For visual realism, we report FID~\cite{heusel2017gans}, which measures differences between deep feature distributions. For text rendering correctness, we report word accuracy (ACC) and normalized edit distance (NED), computed with the official text recognizer~\cite{whatwrong}. We also evaluate the recognizer on the target images.

\begin{table}[ht]
\caption{\textbf{ScenePair results.} \textbf{Boldface} indicates the best results for each metric, while \underline{underlines} highlight the second best results.}
\centering
\begin{tabular}{lccc}
\toprule
\textbf{Method} & \textbf{ACC (\%)$\uparrow$} & \textbf{NED$\uparrow$} & \textbf{FID$\downarrow$} \\
\midrule
SRNet~\citeyear{wu2019editing} & 17.84 & 0.478 & 49.22 \\
MOSTEL~\citeyear{mostel} & 37.69 & 0.557 & 49.19 \\
AnyText~\citeyear{anytext} & 51.12 & 0.734 & 51.79 \\
TextCtrl~\citeyear{zeng2024textctrl} & 84.67 & 0.936 & 43.78 \\
TextSSR~\citeyear{textssr} & 79.69 & 0.924 & 45.37 \\
RS-STE~\citeyear{rsste} & \textbf{91.80} & \textbf{0.981} & 49.59 \\
Flux.1 Kontext~\citeyear{flux} & 12.81 & 0.460 & \underline{41.59} \\
Qwen-Image-Edit~\citeyear{wu2025qwen} & 53.16 & 0.825 & 49.91 \\
TextFlux~\citeyear{textflux} & 80.40 & 0.911 & 54.64 \\
TextRIRO & \underline{85.23}	& \underline{0.964} &	\textbf{41.53} \\
\midrule
\textit{Real (upper bound)} & 88.44 & 0.974 & 0.00 \\
\bottomrule
\end{tabular}
\label{tab:scene_edit_comparison}
\end{table}

The quantitative results in Tab.~\ref{tab:scene_edit_comparison} reveal that early GAN-based methods underperform both in accuracy and visual realism. When diffusion methods were introduced to the STE task, both metrics improve substantially. However, because these methods rely on synthetic data, a gap to real-data performance remains. TextRIRO leverages a self-paired real data training paradigm, achieving near-realistic accuracy and fidelity. Notably, RS-STE achieves higher text recognition accuracy, even surpassing the ground truth, but exhibits poorer FID scores, indicating that it trades realism for correctness and produces strongly synthetic-looking samples. General-purpose editing models complete part of the task, but they lag behind specialized editors, suggesting that expert STE models remain valuable.

Fig.~\ref{fig:comp_vis} presents qualitative visual comparisons: TextRIRO clearly outperforms other methods in edit correctness, style preservation, and realism. MOSTEL, due to its two-stage editing process and training on synthetic data, does not even recognize the existence of these texts, let alone erase them. Even when it attempts to edit, it leaves residual artifacts, as in the ``Buses'' example. TextCtrl uses implicit disentanglement: it keeps coarse structure during feature mapping but loses fine-grained style cues, so the edited text deviates from the original style. RS-STE's samples further highlight its lack of visual authenticity and synthetic appearance, particularly for the ``open'' example. TextSSR, as an inpainting method, depends strongly on surrounding context: when nearby text shares a similar style (e.g., ``open''), it works well, but without such references (e.g., ``51''), its edits degrade. Among general-purpose editors, Flux.1 Kontext preserves the image style but rarely produces the correct target text. Qwen-Image-Edit understands the editing instruction, yet its outputs look rigid and over-synthesized, resembling results from a rendering tool.

\begin{table}[ht]
\footnotesize
\caption{\textbf{RIROCommon results.} Comparison of TextRIRO with three other representative STE models. ``E (Equal)'' denotes text editing with equal-length substitutions, while ``V (Variable)'' corresponds to editing without text length limitations.}
  \centering
  \setlength\tabcolsep{2.5pt}
  \begin{tabular}{l|cc|cc|cc|cc|cc|cc|cc}
  \toprule
\multirow{2}{*}{\textbf{Method}} & \multicolumn{2}{c|}{\textbf{IIIT}~\citeyear{IIIT5K}} & \multicolumn{2}{c|}{\textbf{SVT}~\citeyear{Wang2011SVT}} & \multicolumn{2}{c}{\textbf{IC13}~\citeyear{icdar2013}} & \multicolumn{2}{c|}{\textbf{IC15}~\citeyear{icdar2015}} & \multicolumn{2}{c|}{\textbf{SVTP}~\citeyear{SVTP}} & \multicolumn{2}{c|}{\textbf{CUTE}~\citeyear{Risnumawan2014cute}} & \multicolumn{2}{c}{\textbf{Avg}}  \\
   & E & V & E & V & E & V & E & V & E & V & E & V & E & V \\
  \midrule
  MOSTEL \citeyear{mostel} & 16.85 & 12.85 & 10.82 & 3.58 & 10.04 & 7.54 & 11.49 & 6.15 & 7.91 & 3.98 & 7.89 & 5.02 & 10.83 & 6.52 \\
  TextCtrl \citeyear{zeng2024textctrl} & 64.42 & 11.31 & 76.97 & 10.17 & 74.45 & 12.99 & 73.16 & 15.65 & 70.85 & 10.44 & 64.16 & 11.30 & 70.67 & 11.98 \\
  RS-STE \citeyear{rsste} & 58.80 & 13.21 & 67.08 & 10.73 & 63.71 & 20.11 & 61.29 & 9.10 & 62.17 & 7.59 & 48.39 & 9.62 & 60.24 & 11.73 \\
  \textbf{TextRIRO (ours)} & \textbf{82.74} & \textbf{50.47} & \textbf{84.39} & \textbf{46.14} & \textbf{86.23} & \textbf{53.49} & \textbf{77.91} & \textbf{45.28} & \textbf{79.69} & \textbf{42.31} & \textbf{64.87} & \textbf{48.12} & \textbf{79.31} & \textbf{47.64} \\
  \bottomrule
  \end{tabular}
\label{tab:exp_commonacc}
\end{table}

\noindent\textbf{Broader Editing Evaluation.}
We observe that ScenePair reveals a recognition accuracy ceiling of 88.44\%, largely limited by the STR model. To obtain more faithful results, we adopt a stronger and third-party text recognizer, SVTRv2~\cite{du2024svtrv2}, for evaluation. TextRIRO, together with TextCtrl and RS-STE, already approaches the realism of natural images in ScenePair. Subsequent experiments therefore focus on these three methods, and we conduct an upgraded evaluation on the Common Benchmarks (hereafter RIROCommon). For each image, we provide two targets to edit: an equal-length text that matches the source length and a variable-length target text. We then report normalized accuracy like STR, where we remove non-vocabulary characters and spaces and lowercase all text before scoring.

Tab.~\ref{tab:exp_commonacc} demonstrates TextRIRO's superior performance on RIROCommon: it consistently outperforms alternative models for both equal-length and variable-length editing. In particular, its variable-length editing achieves an accuracy of 47.64\%, markedly higher than competitors, attributable to the inherent diversity of self-paired training examples produced via Shuffle\&Crop. Conversely, RS-STE's strategy of trading off style for accuracy fails under broader and more challenging conditions. And TextRIRO sustains strong performance, achieving equal-length editing accuracy 19.07\% higher than RS-STE.

\begin{table}[ht]
\caption{\textbf{Train-on-edits STR proxy.} “Regular” indicates the average results on the IIIT, SVT, and IC13 datasets, while “Irregular” denotes the average results on the IC15, SVTP, and CUTE datasets.}
  \centering
  \resizebox{\textwidth}{!}{
  \setlength\tabcolsep{2pt}
  \begin{tabular}{l|c|c|c|c|c|c|c|c|c}
  \toprule
\textbf{Method} & IIIT~\citeyear{IIIT5K} & SVT~\citeyear{Wang2011SVT} & IC13~\citeyear{icdar2013} & \textbf{Regular} & IC15~\citeyear{icdar2015} & SVTP~\citeyear{SVTP} & CUTE~\citeyear{Risnumawan2014cute} & \textbf{Irregular} & \textbf{Avg} \\
  \midrule
  MOSTEL \citeyear{mostel} & 51.87 & 59.51 & 64.99 & 58.79 & 44.62 & 44.34 & 34.72 & 41.22 & 50.01 \\
  AnyText \citeyear{anytext} & 73.27 & 78.83 & 82.61 & 78.24 & 66.08 & 70.46 & 48.27 & 61.60 & 69.92 \\
  TextCtrl \citeyear{zeng2024textctrl} & 79.40 & 84.39 & 89.61 & 84.47 & 72.06 & 77.36 & 53.12 & 67.51 & 75.99 \\
  RS-STE \citeyear{rsste} & 75.43 & 85.32 & 87.98 & 82.91 & 74.43 & 78.45 & 59.72 & 70.87 & 76.89 \\
  \textbf{TextRIRO (ours)} & \textbf{90.53} & \textbf{89.18} & \textbf{92.65} & \textbf{90.79} & \textbf{81.00} & \textbf{84.03} & \textbf{82.64} & \textbf{82.56} & \textbf{86.67} \\
  \midrule
  \textit{Real (upper bound)} & 99.30 & 97.84 & 98.60 & 98.58 & 90.50 & 96.12 & 98.26 & 94.96 & 96.77 \\
  \bottomrule
  \end{tabular}}
\label{tab:exp_trainedit}
\end{table}

\noindent\textbf{STR Training Proxy Task.}
Although we employ a strong recognizer, it cannot reflect absolute editing accuracy. Following TextSSR~\cite{textssr}, we use the edited text to train an STR model and treat it as a proxy evaluation of the realism and accuracy of generated text. Only sufficiently realistic and accurate synthetic data will improve downstream recognition performance. This also serves as a practical measure of whether editing models can effectively generate usable training samples. We apply each STE model to U14M-Filter, editing each image with the same alternative string of equal length, and train STR models with edited images and their edited labels, then evaluate them on the STR protocol.

Tab.~\ref{tab:exp_trainedit} confirms previous trends: MOSTEL's outdated architecture and data strategy result in poor performance. TextCtrl and RS-STE utilize different architectures and training data, yet achieve similar results, with only a 0.90\% difference in average accuracy. Both methods follow training paradigms dominated by synthetic data, revealing the upper bound and bottleneck of such approaches, as the domain gap between synthetic and real data persists despite model optimization. In contrast, TextRIRO's self-paired training paradigm leverages real data to overcome this bottleneck, yielding an average accuracy of 86.67\%, representing a 9.78\% improvement over the next-best approach. This advantage is even more pronounced on irregular samples, with an 11.69\% gap, highlighting TextRIRO's ability to supplement scarce training domains.

\begin{figure}[ht]
\centering
\begin{minipage}{0.9\textwidth}
  \centering
  \begin{subfigure}[c]{0.18\linewidth}
    \centering
    \includegraphics[width=\linewidth]{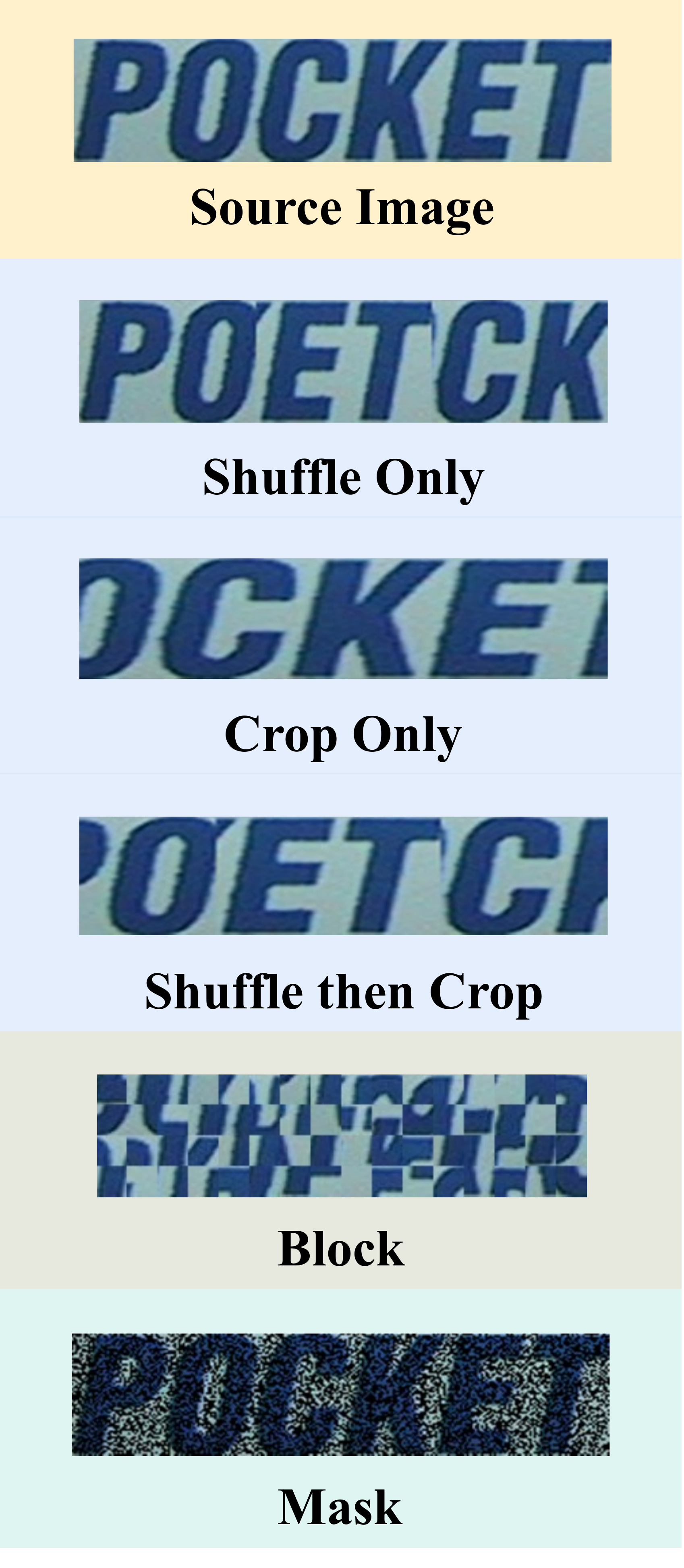}
  \end{subfigure}\hfill%
  \begin{subfigure}[c]{0.72\linewidth}
    \centering
    \setlength{\tabcolsep}{3pt}
    \renewcommand{\arraystretch}{1.12}
    \resizebox{\linewidth}{!}{%
      \begin{tabular}{cccc|ccc}
        \toprule
        \textbf{Text Enc.} &
        \textbf{Shape} &
        \textbf{VAE} &
        \textbf{Paired Op.} &
        \textbf{ACC $\uparrow$} &
        \textbf{NED $\uparrow$} &
        \textbf{FID $\downarrow$} \\
        \midrule
        None  & [256,256] & SD 1.5 & Shuffle\&Crop &
        9.38 & 0.469 & 55.84 \\
        Glyph & [256,256] & SD 1.5 & Shuffle\&Crop &
        53.44 & 0.815 & 40.36 \\
        STR   & [256,256] & SD 1.5 & Shuffle\&Crop &
        56.72 & 0.823 & 39.47 \\
        STR   & [64,256] & SD 1.5 & Shuffle\&Crop &
        78.91 & 0.951 & 54.04 \\
        STR   & [64,256] & SD 2.1 & Shuffle\&Crop &
        \textbf{82.42} & \textbf{0.960} & 43.59 \\
        Glyph & [64,256] & SD 2.1 & Shuffle\&Crop &
        81.95 & 0.952 & 45.96 \\
        \midrule
        STR & [64,256] & SD 2.1 & Shuffle only &
        45.78 & 0.791 & 39.10 \\
        STR & [64,256] & SD 2.1 & Crop only &
        80.08 & 0.954 & 45.46 \\
        STR & [64,256] & SD 2.1 & Mask &
        13.13 & 0.493 & 58.08 \\
        STR & [64,256] & SD 2.1 & Block &
        79.36 & 0.949 & 50.67 \\
        STR & [64,256] & SD 2.1 & None (identity) &
        4.37 & 0.176 & \textbf{33.28} \\
        \bottomrule
      \end{tabular}%
    }
  \end{subfigure}
\end{minipage}
\caption{\textbf{Ablations on ScenePair.}
Left: visualizations of each self-pairing operator.
Right: full factorial over text encoder pre-training, input shape,
VAE variant and paired-data operator.
Shuffle\&Crop strictly dominates every other operator when joined
with STR-pretrained text conditioning and a rectangular input.
``None'' collapses to an identity mapping (no perturbation operation).}
\label{fig:ablation}
\end{figure}

\subsection{Ablation Study}
\label{sec:ablation}

\begin{figure*}[h]
\centering
\includegraphics[width=\textwidth]{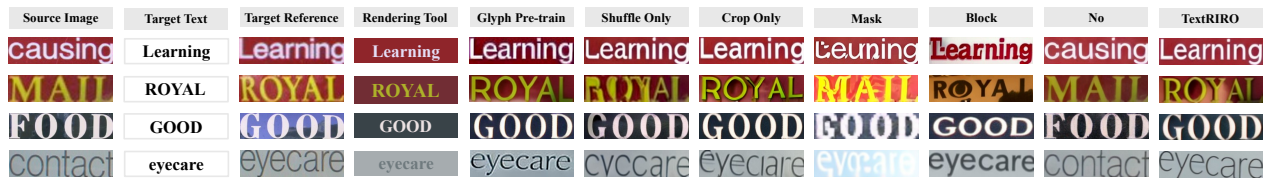} 
\caption{\textbf{Qualitative ablations.} The showing setting is consistent with that in Fig.~\ref{fig:comp_vis}.}
\label{fig:comp_vis_abl}
\end{figure*}

\noindent\textbf{Contributions of each core component.} Fig.~\ref{fig:ablation} summarizes ablations over the TextRIRO architecture and self-paired training paradigm. Text encoding is essential: without pre-training, the model fails to learn reliable character correspondences, while STR pre-training performs better than glyph pre-training because it is learned directly from real scene text. SD~2.1 provides better VAE reconstruction than SD~1.5, and rectangular $64\times256$ inputs avoid the distortion caused by square resizing. For self-paired data construction, Shuffle\&Crop achieves the best accuracy/realism trade-off. Crop Only lacks diversity, Shuffle Only breaks alignment, Mask destroys both content and style, Block introduces a strong distribution shift, and removing paired construction degenerates to identity copying. Fig.~\ref{fig:comp_vis_abl} visualizes key ablation findings. The Glyph Pre-train carries strong glyph priors and tends to lose style details, while the STR Pre-train better maintains the original style during editing. Alternative self-paired constructions also trail Shuffle\&Crop in both rendering accuracy and style fidelity: Shuffle Only and Crop Only give less diversity, and Mask and Block create a large train–test gap.

\noindent\textbf{Layout-wise attribution of Shuffle and Crop.}
To make the complementary roles of the two perturbations explicit, we add an attribution study (in Tab.~\ref{tab:layout_attribution}) on straight (IC13), curved (CUTE), and multi-oriented (SVTP) subsets, separately running Shuffle only, Crop only, and Shuffle\&Crop. The reading of our design is as follows. Shuffle dominates for straight text, because it maximally breaks left-to-right semantics while preserving column-local style. Crop dominates for curved and multi-oriented text, because it preserves local tilt and curvature within contiguous sub-images.

\begin{table}[ht]
\caption{Layout-wise attribution of different operators.  Values are
equal-length editing accuracy (\%).}
\centering
\begin{tabular}{lccc}
\toprule
\textbf{Operator} & \textbf{Straight (IC13)} & \textbf{Curved (CUTE)} & \textbf{Multi-oriented (SVTP)} \\
\midrule
Shuffle only & 79.00 & 54.83 & 66.67 \\
Crop only & 77.83 & \textbf{60.93} & \textbf{69.76} \\
Shuffle\&Crop & \textbf{80.16} & 58.42 & 68.84 \\
\bottomrule
\end{tabular}
\label{tab:layout_attribution}
\end{table}

\noindent\textbf{DiT-Based Variant.} We also try to instantiate TextRIRO with DiT-based conditional control, using OminiControl~\cite{tan2025ominicontrol} as a representative lightweight condition-injection method. This variant can be trained under our self-paired real-data formulation, but its ScenePair performance remains behind the ControlNet-based TextRIRO. The main reason is that OminiControl-style conditioning injects image conditions in a compact token-level manner, which is parameter-efficient and general, but relatively coarse for scene text. For text editing, small local deviations can change character identity or introduce artifacts. Therefore, ControlNet's multi-scale feature injection is better suited because it preserves and transfers local style cues at several spatial resolutions. In addition, the OminiControl paper reports that, despite its minimal and universal design, it is not uniformly superior to ControlNet across all controlled generation settings. We therefore use ControlNet as the default backbone and leave more specialized DiT-based fine-grained control as future work.

\begin{table}[ht]
\caption{Backbone choice comparison on ScenePair~\cite{zeng2024textctrl}.}
\centering
\begin{tabular}{lccc}
\toprule
\textbf{Backbone} & \textbf{ACC (\%)$\uparrow$} & \textbf{NED$\uparrow$} & \textbf{FID$\downarrow$} \\
\midrule
OminiControl-based~\citeyear{tan2025ominicontrol} & 52.89 & 0.803 & \textbf{39.65} \\
ControlNet-based~\citeyear{controlnet} & \textbf{85.23} & \textbf{0.964} & 41.53 \\
\bottomrule
\end{tabular}
\label{tab:control_backbone}
\end{table}

\noindent\textbf{Cross-recognizer evaluation.}
ScenePair ACC is measured with TRBA~\cite{whatwrong} as Official.  To
test whether the observed accuracy--realism trade-off is recognizer-specific,
we additionally evaluate with SVTRv2~\cite{du2024svtrv2} and PPOCR-v6-rec~\cite{zhang2026pp}.  Tab.~\ref{tab:cross_recognizer}
shows the same trend for all three recognizers: TextRIRO remains close to the
real-image upper bound, whereas RS-STE consistently exceeds it while having a
substantially worse FID (49.59 versus 41.53).  This supports the interpretation
that RS-STE produces recognizer-friendly but less realistic images.

\begin{table}[ht]
\caption{ScenePair word accuracy (\%) with three independent recognizers.}
\centering
\begin{tabular}{lccc}
\toprule
\textbf{Method} & \textbf{Official} & \textbf{SVTRv2} & \textbf{PPOCR-v6-rec} \\
\midrule
MOSTEL~\citeyear{mostel} & 37.69 & 33.91 & 37.42 \\
TextCtrl~\citeyear{zeng2024textctrl} & 84.67 & 85.08 & 88.67 \\
TextSSR~\citeyear{textssr} & 79.69 & 80.23 & 82.34 \\
RS-STE~\citeyear{rsste} & \textbf{91.80} & \textbf{90.39} & \textbf{91.80} \\
TextRIRO & 85.23 & 85.86 & 86.25 \\
Real (upper bound) & 88.44 & 88.67 & 89.14 \\
\bottomrule
\end{tabular}
\label{tab:cross_recognizer}
\end{table}

\subsection{Synthetic Data Utility}

Despite TextRIRO's advances, a 10.10\% accuracy gap remains compared to training with the same amount of real data. This gap mainly comes from a non-trivial number of edit failures that slip into the training set and accumulate errors. TextSSR~\cite{textssr} filters with an STR model to select usable samples, but the retained images largely duplicate styles already present in real data, which causes distribution redundancy and limits data expansion for STR. Because STE generates text that closely matches the original style, we concatenate each original image with its filtered edited result to build TextRIRO-3M, a long-text synthetic dataset.

\begin{table}[ht]
  \caption{\textbf{Data quality.} Performance comparison of the STR model (MAERec~\cite{jiang2023revisiting}) with different types of synthetic data mixed with real data.}
  \centering
  \resizebox{\textwidth}{!}{
  \setlength\tabcolsep{2pt}
  \begin{tabular}{l|c|ccccccc|cccccccc}
  \toprule
  \multirow{2}{*}{\textbf{Training Data}} & \multirow{2}{*}{\textbf{Volume}} & \multicolumn{7}{c|}{\textbf{Common Benchmarks}} & \multicolumn{8}{c}{\textbf{Union14M-Benchmark}} \\
   & &
  \textbf{IIIT} & \textbf{SVT} & \textbf{IC13} & \textbf{IC15} & \textbf{SVTP} & \textbf{CUTE} & \textbf{Avg} &
  \textbf{CUR} & \textbf{MLO} & \textbf{ART} & \textbf{CTL} & \textbf{SAL} & \textbf{MLW} & \textbf{GEN} & \textbf{Avg} 
  \\
  \midrule
  U14M-Filter~\citeyear{du2024svtrv2} (UF) & 3.22M & 99.30 & \textbf{97.84} & 98.60 & 90.50 & 96.12 & 98.26 & 96.77 & 89.49 & 93.50 & 77.11 & 83.18 & 86.99 & 86.89 & 85.48 & 86.09 \\
  MJ~\citeyear{mj}~+~UF & 1M~+~3.22M & 99.17 & 97.53 & \textbf{99.18} & 91.44 & 95.81 & 98.26 & 96.89 & 90.35 & 94.01 & 79.11 & 85.62 & 87.3 & 87.26 & 85.84 & 87.07 \\
  ST~\citeyear{st}~+~UF & 1M~+~3.22M & 99.33 & 97.53 & 98.95 & 90.94 & 96.74 & 98.26 & 96.96 & 90.44 & 94.08 & 78.00 & 86.14 & \textbf{87.81} & 87.01 & 85.77 & 87.04 \\
  TextSSR~\citeyear{textssr}~+~UF & 1M~+~3.22M & 99.07 & 97.53 & 98.83 & 91.28 & 95.50 & 98.61 & 96.80 & 91.06 & 94.52 & \textbf{81.22} & 86.26 & 87.56 & 88.35 & 85.95 & 87.85 \\
  \midrule
  MOSTEL~\citeyear{mostel}~+~UF & 1M~+~3.22M & 99.17 & 97.53 & 98.37 & 91.00 & \textbf{96.90} & 96.87 & 96.64 & 90.89 & 93.64 & 79.33 & 83.83 & 87.43 & 87.50 & 85.74 & 86.91 \\
  TextCtrl~\citeyear{zeng2024textctrl}~+~UF & 1M~+~3.22M & 99.17 & \textbf{97.84} & 98.48 & \textbf{91.61} & 95.81 & 98.26 & 96.86 & 90.35 & 94.01 & 79.11 & 84.34 & 87.30 & 87.26 & 85.84 & 86.89 \\
  RS-STE~\citeyear{rsste}~+~UF & 1M~+~3.22M & 99.10 & 97.06 & 98.6 & 91.39 & 96.74 & \textbf{99.65} & 97.09 & 90.89 & 94.16 & 80.67 & 85.75 & 87.49 & 86.77 & 85.80 & 87.36 \\
  TextRIRO~+~UF & 1M~+~3.22M & \textbf{99.37} & \textbf{97.84} & 98.72 & 91.33 & 96.59 & 98.96 & \textbf{97.13} & \textbf{91.18} & \textbf{94.67} & 81.00 & \textbf{86.65} & 87.49 & \textbf{89.08} & 85.80 & \textbf{87.98} \\
  Real-Concat~+~UF & 1M~+~3.22M & 99.27 & 97.53 & 98.95 & 90.50 & 95.97 & 98.61 & 96.81 & 90.89 & 94.59 & 79.67 & 85.24 & 86.73 & 88.23 & \textbf{86.01} & 87.33 \\
  \bottomrule
  \end{tabular}}
  \label{tab:comp_data}
\end{table}

We mix each 1M synthetic dataset with the real dataset to compare how different synthetic sources improve recognition. The candidates include MJ~\cite{mj} and ST~\cite{st}, the diffusion-based TextSSR~\cite{textssr}, and datasets produced by the same concatenation strategy with MOSTEL~\cite{mostel}, TextCtrl~\cite{zeng2024textctrl}, and RS-STE~\cite{rsste}, which we compare against TextRIRO. Tab.~\ref{tab:comp_data} shows that TextRIRO yields the best improvements on both the standard Common Benchmarks and the more challenging Union14M-Benchmark. On the latter, the Multi-words subset improves by 2.19\%, which aligns with the benefit of longer text introduced by concatenation. Although other editing-based generators also adopt concatenation, they remain bounded by the synthesis ceiling discussed above, and their performance still falls short of TextRIRO. Furthermore, we concatenate two independent real images to create longer sequences (Real-Concat). This yields some STR improvements, confirming that sequence-length augmentation helps. However, the performance is still lower than with TextRIRO-generated data. We attribute this gap to style inconsistency between concatenated samples.

\newcommand{\deltaval}[1]{{\small #1}}
\begin{table}[ht]
\caption{Extension of the experiments in Tab.~\ref{tab:comp_data}, showing the performance of TextRIRO synthetic data on various STR models and data scales.}
\centering
\resizebox{\textwidth}{!}{
\setlength\tabcolsep{2pt}
\begin{tabular}{l|c|c|ccccccc|cccccccc}
\toprule
\multirow{2}{*}{\textbf{Model}} &
\multirow{2}{*}{\textbf{Training Data}} &
\multirow{2}{*}{\textbf{Volume}} &
\multicolumn{7}{c|}{\textbf{Common Benchmarks}} &
\multicolumn{8}{c}{\textbf{Union14M-Benchmark}} \\
& & &
\textbf{IIIT} & \textbf{SVT} & \textbf{IC13} &
\textbf{IC15} & \textbf{SVTP} & \textbf{CUTE} & \textbf{Avg} &
\textbf{CUR} & \textbf{MLO} & \textbf{ART} & \textbf{CTL} &
\textbf{SAL} & \textbf{MLW} & \textbf{GEN} & \textbf{Avg} \\
\midrule

\multirow{5}{*}{MAERec~\citeyear{jiang2023revisiting}} &
UF & 3.22M &
99.30 & \textbf{97.84} & 98.60 & 90.50 & 96.12 & 98.26 & 96.77 &
89.49 & 93.50 & 77.11 & 83.18 & 86.99 & 86.89 & 85.48 & 86.09 \\

& \multirow{2}{*}{TextRIRO~+~UF} &
\multirow{2}{*}{1M~+~3.22M} &
\textbf{99.37} & \textbf{97.84} & \textbf{98.72} & 91.33 &
\textbf{96.59} & \textbf{98.96} & \textbf{97.13} &
91.18 & \textbf{94.67} & \textbf{81.00} & \textbf{86.65} &
87.49 & 89.08 & 85.80 & 87.98 \\

& & &
\deltaval{\textcolor{blue}{+0.07}} &
\deltaval{+0.00} &
\deltaval{\textcolor{blue}{+0.12}} &
\deltaval{\textcolor{blue}{+0.83}} &
\deltaval{\textcolor{blue}{+0.47}} &
\deltaval{\textcolor{blue}{+0.70}} &
\deltaval{\textcolor{blue}{+0.36}} &
\deltaval{\textcolor{blue}{+1.69}} &
\deltaval{\textcolor{blue}{+1.17}} &
\deltaval{\textcolor{blue}{+3.89}} &
\deltaval{\textcolor{blue}{+3.47}} &
\deltaval{\textcolor{blue}{+0.50}} &
\deltaval{\textcolor{blue}{+2.19}} &
\deltaval{\textcolor{blue}{+0.32}} &
\deltaval{\textcolor{blue}{+1.89}} \\

& \multirow{2}{*}{TextRIRO~+~UF} &
\multirow{2}{*}{3M~+~3.22M} &
99.20 & \textbf{97.84} & \textbf{98.72} & \textbf{91.50} &
95.66 & \textbf{98.96} & 96.98 &
91.92 & 94.38 & 80.89 & 84.85 &
\textbf{87.74} & \textbf{90.17} & 86.07 & \textbf{88.00} \\

& & &
\deltaval{\textcolor{red}{-0.10}} &
\deltaval{+0.00} &
\deltaval{\textcolor{blue}{+0.12}} &
\deltaval{\textcolor{blue}{+1.00}} &
\deltaval{\textcolor{red}{-0.46}} &
\deltaval{\textcolor{blue}{+0.70}} &
\deltaval{\textcolor{blue}{+0.21}} &
\deltaval{\textcolor{blue}{+2.43}} &
\deltaval{\textcolor{blue}{+0.88}} &
\deltaval{\textcolor{blue}{+3.78}} &
\deltaval{\textcolor{blue}{+1.67}} &
\deltaval{\textcolor{blue}{+0.75}} &
\deltaval{\textcolor{blue}{+3.28}} &
\deltaval{\textcolor{blue}{+0.59}} &
\deltaval{\textcolor{blue}{+1.91}} \\
\midrule

\multirow{5}{*}{SVTRv2-S~\citeyear{du2024svtrv2}} &
UF & 3.22M &
99.00 & 97.22 & 98.37 & 90.39 & 93.80 & 98.96 & 96.29 &
86.97 & 90.50 & 77.33 & 85.24 & 82.50 & 84.47 &
\textbf{84.28} & 84.47 \\

& \multirow{2}{*}{TextRIRO~+~UF} &
\multirow{2}{*}{1M~+~3.22M} &
99.00 & 97.84 & 98.72 & 90.34 & 94.11 & \textbf{99.31} & 96.55 &
88.09 & 91.67 & 77.67 & 85.24 & \textbf{84.31} &
86.29 & 83.72 & 85.28 \\

& & &
\deltaval{+0.00} &
\deltaval{\textcolor{blue}{+0.62}} &
\deltaval{\textcolor{blue}{+0.35}} &
\deltaval{\textcolor{red}{-0.05}} &
\deltaval{\textcolor{blue}{+0.31}} &
\deltaval{\textcolor{blue}{+0.35}} &
\deltaval{\textcolor{blue}{+0.26}} &
\deltaval{\textcolor{blue}{+1.12}} &
\deltaval{\textcolor{blue}{+1.17}} &
\deltaval{\textcolor{blue}{+0.34}} &
\deltaval{+0.00} &
\deltaval{\textcolor{blue}{+1.81}} &
\deltaval{\textcolor{blue}{+1.82}} &
\deltaval{\textcolor{red}{-0.56}} &
\deltaval{\textcolor{blue}{+0.81}} \\

& \multirow{2}{*}{TextRIRO~+~UF} &
\multirow{2}{*}{3M~+~3.22M} &
\textbf{99.27} & \textbf{97.99} & \textbf{98.83} & \textbf{90.89} &
\textbf{94.73} & 98.61 & \textbf{96.72} &
\textbf{89.65} & \textbf{92.26} & \textbf{78.22} & \textbf{85.49} &
84.09 & \textbf{86.77} & 84.20 & \textbf{85.81} \\

& & &
\deltaval{\textcolor{blue}{+0.27}} &
\deltaval{\textcolor{blue}{+0.77}} &
\deltaval{\textcolor{blue}{+0.46}} &
\deltaval{\textcolor{blue}{+0.50}} &
\deltaval{\textcolor{blue}{+0.93}} &
\deltaval{\textcolor{red}{-0.35}} &
\deltaval{\textcolor{blue}{+0.43}} &
\deltaval{\textcolor{blue}{+2.68}} &
\deltaval{\textcolor{blue}{+1.76}} &
\deltaval{\textcolor{blue}{+0.89}} &
\deltaval{\textcolor{blue}{+0.25}} &
\deltaval{\textcolor{blue}{+1.59}} &
\deltaval{\textcolor{blue}{+2.30}} &
\deltaval{\textcolor{red}{-0.08}} &
\deltaval{\textcolor{blue}{+1.34}} \\
\bottomrule
\end{tabular}
}
\label{tab:comp_data_model}
\end{table}
Tab.~\ref{tab:comp_data_model} assesses generality across models and scalability with data size. With SVTRv2-S~\cite{du2024svtrv2}, TextRIRO still brings consistent gains, especially on Multi-Words samples. Scaling TextRIRO to 3M further improves Union14M performance for MAERec and raises Multi-Words to 90.17\%, although gains on some common subsets saturate.

\section{Conclusion}

In this paper, we have introduced TextRIRO, a scene text editing method that, for the first time, learns editing purely from real data through a self-paired supervision scheme. By disrupting semantic content via targeted shuffle and crop operations while maintaining style, we eliminated dependence on synthetic triplets and bypassed the need for explicit or implicit disentanglement modules. Our diffusion-based model achieves near-realistic fidelity and robust content accuracy, effectively narrowing the synthetic-to-real gap that limits previous STE approaches. Comprehensive experiments on multiple benchmarks demonstrate that TextRIRO surpasses existing STE methods in both correctness and visual realism. Additionally, we create a large-scale, realistic long-text recognition dataset, TextRIRO-3M. Training STR models on it yielded significant gains, proving that high-realism STE continues to propel STR performance beyond current saturation points. Naturally, TextRIRO excels in English thanks to abundant real scene text data, but such multilingual data is scarce. Future work will focus on accurate and realistic multilingual STE in low-resource settings.


\section*{Acknowledgments and Disclosure of Funding}
This work was supported by National Natural Science Foundation of China under Grant 625B2057.

\bibliographystyle{plainnat}
\bibliography{main}

\section*{Appendix}


\begin{figure}[h]  
    \begin{minipage}{0.58\columnwidth}
         \centering
         \includegraphics[width=\linewidth]{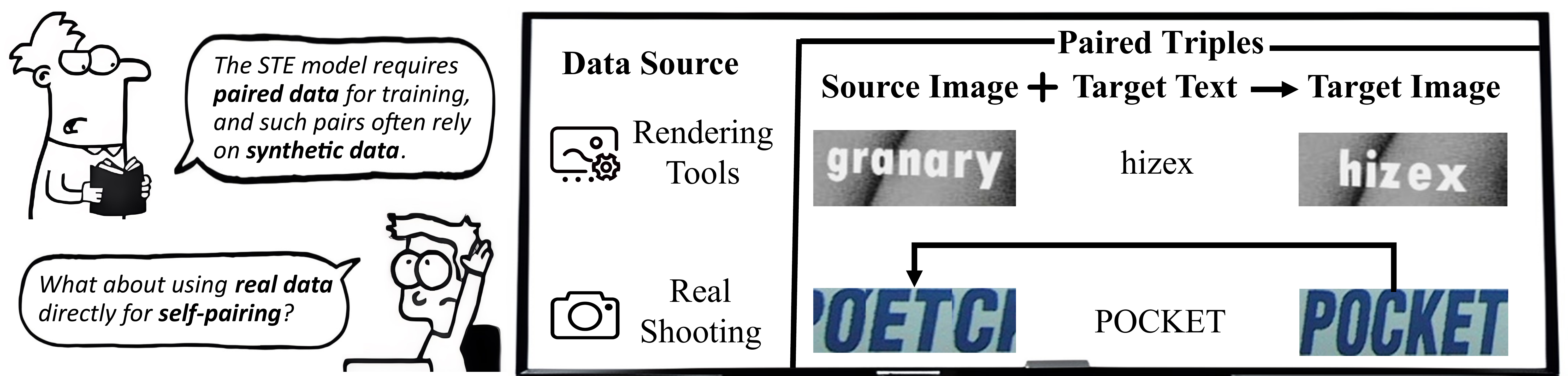}
    \end{minipage}
    \hfill  
    \begin{minipage}{0.4\columnwidth}
         \centering
         \includegraphics[width=\linewidth]{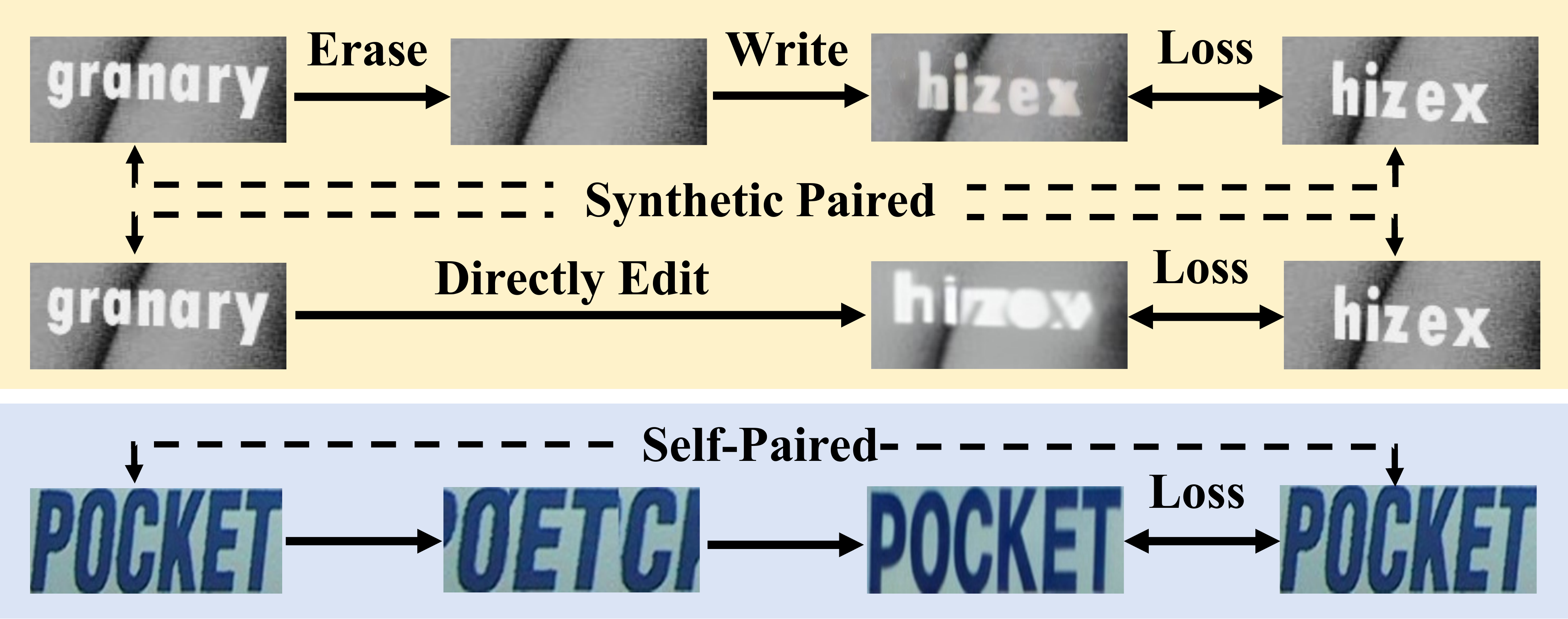}
    \end{minipage}
         \caption{An interesting comic for introducing our idea. \colorbox[HTML]{FEF2CA}{\textbf{Top:}} Two previous STE training paradigms, both requiring paired synthetic data.
        \colorbox[HTML]{DBE4F5}{\textbf{Bottom:}} Our method leverages self-paired real data without any synthetic pairs.}
\label{fig:teaser}
\end{figure}

\section{More Implementation Details}

\subsection{Model Formulation Details}
\label{app:model_formulation}

The details of Sec.~\ref{sec_3-2} are as follows. Given $I_{\text{src}}$, the style encoder first extracts ViT features
\begin{equation}
 F_{\text{vit}} = \text{ViT}(I_{\text{src}}) \in \mathbb{R}^{N \times D},
\end{equation}
where $N$ is the number of patches and $D$ is the feature dimension. For each pyramid level $k$, adaptive average pooling and a zero convolution~\cite{controlnet} produce
\begin{equation}
 F^{(k)}_{\text{pyramid}} = \text{Conv}_{1\times1}(\text{AvgPool}_{(h_k,w_k)}(F_{\text{vit}})),
\end{equation}
and all levels are aggregated as
\begin{equation}
 F_{\text{style}} = \{ F^{(k)}_{\text{pyramid}} \mid k=1,...,K \}.
\end{equation}
The ViT takes RGB images of size $64 \times 256$, uses $16 \times 16$ patches, and outputs 768-dimensional patch embeddings. It contains 12 layers with 12 attention heads. We set $K=4$, with pyramid output sizes $[8,32]$, $[4,16]$, $[2,8]$, and $[1,4]$.

For the text encoder, each target text $T_{\text{tgt}}$ with length $l = |T_{\text{tgt}}|$ is truncated or padded to the maximal character length $L_M$. Each character $t_i$ is mapped to a vocabulary index $x_i$, yielding
\begin{align}
X_{\text{tgt}} &= \langle\mathrm{s}\rangle + [x_1 \ldots x_l] + \langle\mathrm{/s}\rangle + \langle\mathrm{pad}\rangle_{l+3}^{L_M+2},
\end{align}
where $\langle\mathrm{s}\rangle$, $\langle\mathrm{/s}\rangle$, and $\langle\mathrm{pad}\rangle$ are special tokens. The encoded feature is
\begin{equation}
 F_{\text{text}} = \text{Embed}(X_{\text{tgt}}) + \text{PosEnc}(X_{\text{tgt}})\in \mathbb{R}^{ (L_M+2) \times D}.
\end{equation}
We set $L_M=25$ and use 94 printable characters plus the space character. The embedding dimension is 768, and the text feature is fed to the conditional U-Net through cross-attention. The conditional diffusion loss is
\begin{equation}
 \mathcal{L}_\text{TextRIRO} = \| \epsilon - \hat{\epsilon}(z_t, t, F_{\text{style}}, F_{\text{text}}) \|^2,
\end{equation}
where $z_t$ is the noisy latent at timestep $t$, $\epsilon$ is the injected noise, and $\hat{\epsilon}$ is the predicted noise conditioned on style and text features.

\subsection{Model specific settings}

In the setup of Sec.~\ref{sec_3-1} and Alg.~\ref{alg_data}, we set $p_\text{shuffle}$ to $0.7$ by default. To investigate the impact of this parameter, we conduct a rough ablation study with values $0.5$, $0.7$, and $1.0$. The results are summarized in Tab.~\ref{tab:shuffle_ablation}. We observe that setting $p_\text{shuffle}$ to $0.7$ yields the best overall results in terms of accuracy and NED, while maintaining competitive FID. Lower values result in reduced accuracy, whereas higher values slightly degrade FID without substantial gains in other metrics. This suggests that $p_\text{shuffle}=0.7$ achieves a balanced trade-off among all evaluated criteria, and we thus adopt it as our default choice for subsequent experiments.

\begin{table}[ht]
\caption{Ablation study of $p_\text{shuffle}$ on ScenePair~\cite{zeng2024textctrl}.}
\centering
\begin{tabular}{lccc}
\toprule
\textbf{$p_\text{shuffle}$} & \textbf{ACC (\%)$\uparrow$} & \textbf{NED$\uparrow$} & \textbf{FID$\downarrow$} \\
\midrule
0.5 & 80.78 & 0.943 & \textbf{42.76} \\
0.7 & \textbf{82.42} & \textbf{0.960} & 43.59 \\
1.0 & 81.72 & 0.947 & 44.01 \\
\bottomrule
\end{tabular}
\label{tab:shuffle_ablation}
\end{table}



\subsection{Datasets and Implementation Details}

\noindent\textbf{Training Datasets.}
Our self-paired pipeline uses the text label and, in particular, its length,
but it does \emph{not} require character segmentation.  For Shuffle, we split
the image width into $n_{\mathrm{split}}\in[2,L]$ equal-width columns, where
$L$ is the label length, and randomly permute the columns.  Crop takes a
horizontal contiguous sub-interval.  Thus, a character may be cut by a column
boundary. This is intentional because the purpose is to remove readable
semantics while retaining local visual style.  We train our STE model and the
STR baseline on Union14M~\cite{jiang2023revisiting}, currently the largest
available scene text dataset. Specifically, we adopt a filtered version of its
labeled subset, denoted as U14M-Filter~\cite{du2024svtrv2}, which contains
3.22M annotated text instances. Note that training in a completely label-free
manner is theoretically infeasible, because the definition of STE inherently
requires specifying the edited text.

\noindent\textbf{Evaluation Datasets.}
To evaluate scene text editing performance under real-world conditions, we first employ the ScenePair~\cite{zeng2024textctrl}, which comprises 1,280 pairs of carefully collected, real-world text images. Furthermore, we construct benchmarks for equal-length and variable-length text editing using six widely-used text recognition datasets: IIIT 5K-Words (IIIT~\cite{IIIT5K}), ICDAR2013 (IC13~\cite{icdar2013}), Street View Text (SVT~\cite{Wang2011SVT}), ICDAR2015 (IC15~\cite{icdar2015}), Street View Text-Perspective (SVTP~\cite{SVTP}), and CUTE80 (CUTE~\cite{Risnumawan2014cute}), which we collectively refer to as the Common Benchmarks. Additionally, we assess data quality using Union14M-Benchmark~\cite{jiang2023revisiting}, which includes seven subsets: Curve (CUR), Multi-Oriented (MLO), Artistic (ART), Contextless (CTL), Salient (SAL), Multi-Words (MLW), and General (GEN).

\noindent\textbf{Implementation Details.}
We use MAERec~\cite{jiang2023revisiting} both for text-encoder pre-training and as our STR proxy baseline, training for 20 epochs under the settings established by the OpenOCR framework~\cite{du2024svtrv2}. TextRIRO builds upon SD~2.1 for the VAE and SD~1.5 for the U-Net backbone~\cite{sd}. The extensively pre-trained VAE in SD~2.1 is well-suited for scene text tasks. The U-Net requires its parameters to be unfrozen for training, so the specific version has minimal impact. Meanwhile, SD~1.5 uses a smaller dimension ($\mathrm{dim}=768$) and can directly match the feature dimension of the text encoder. All images are resized to $64 \times 256$ and those with aspect ratios less than $1\!:\!2$ are rotated $90^\circ$ counterclockwise to match the target resolution. We use a batch size of 256, apply an initial learning rate of $1\times10^{-5}$, and train for 15 epochs on 8 NVIDIA V100 GPUs. During inference, we use 20 sampling steps and a classifier-free guidance (CFG) scale of $\omega=2.2$. For the generation of the TextRIRO-3M dataset, we use only reference images whose original text contains at least 4 characters. Consequently, the minimum label length in TextRIRO-3M is 8. For the other STE models, we use their officially reported results when available. Otherwise, we use their released official checkpoints and the default inference hyperparameters to ensure a fair comparison.

\noindent\textbf{Training--inference protocol.}
During training, the style condition is the perturbed view of the same real
image and the target text is its original label.  At inference, the source is
an unperturbed real image and the target is a new string.  This is an
unavoidable residual mismatch because coherent real source--target pairs with
different text are not available at scale.  Shuffle\&Crop is designed to
reduce the mismatch by producing a real-image-derived style condition whose
semantics are destroyed but whose local appearance remains realistic.  The
self-paired objective also provides a content-disentanglement constraint: the
source condition has disrupted text order, while the target text is supplied
only through the text encoder.  If semantic content leaked through the style
path, the model would copy the disrupted content rather than render the
correctly ordered target.  Consistently, the identity (None) ablation in
Fig.~\ref{fig:ablation} collapses to an ACC of 4.37, showing that an
unperturbed source encourages copying instead of text-conditioned editing.

\noindent\textbf{Computational Overhead and Runtime Efficiency.} We report the computational requirements for both the STR pre-training and U-Net fine-tuning stages, which use 36 GPU-hours and 180 GPU-hours, respectively. For inference, we evaluate our method on a single NVIDIA V100 GPU with \texttt{diffusion\_steps} set to 20 and a batch size of 32. Processing 10{,}000 batches requires 18 hours, corresponding to an average inference time of 0.20 seconds per image.

\begin{table}[ht]
\caption{Performance of different VAE variants on the quality of reconstructed scene text images.}
\resizebox{\textwidth}{!}{
  \centering
  \setlength\tabcolsep{2pt}
{
\begin{tabular}{lc|ccccccc} 
\toprule
\textbf{Metrics} & \textbf{VAE Version} & \textbf{IIIT}~\citeyear{IIIT5K} & \textbf{SVT}~\citeyear{Wang2011SVT} & \textbf{IC13}~\citeyear{icdar2013} & \textbf{IC15}~\citeyear{icdar2015} & \textbf{SVTP}~\citeyear{SVTP} & \textbf{CUTE}~\citeyear{Risnumawan2014cute} & \textbf{Avg} 
\\  
    \midrule
    \multirow{3}{*}{\textbf{PSNR$\uparrow$}} & SD~1.5    & 25.90 & 30.79 & 28.41 & 29.87 & 28.58 & 26.43 & 28.33 \\
  &  SD~2.1    & 27.63 & \textbf{33.02} & 30.23 & 31.99 & 30.68 & 28.30 & 30.31 \\
   & SD~3.5    & \textbf{29.26} & 32.34 & \textbf{30.32} & \textbf{32.18} & \textbf{31.64} & \textbf{28.87} & \textbf{30.77} \\
    \midrule
    \multirow{3}{*}{\textbf{FID$\downarrow$}} &
    SD~1.5    & 1.36 & 0.50 & 0.50 & 1.22 & 1.83 & 0.50 & 0.98 \\
   & SD~2.1    & \textbf{0.75} & \textbf{0.21} & \textbf{0.24} & \textbf{0.71} & \textbf{1.03} & \textbf{0.16} & \textbf{0.52} \\
   & SD~3.5    & 1.28 & 0.39 & 0.41 & 1.14 & 1.72 & 0.34 & 0.88 \\
    \midrule
\multirow{4}{*}{\textbf{Acc (\%)$\uparrow$}} &
    SD~1.5    & 98.70 & 97.37 & 97.67 & 89.62 & 92.56 & 97.57 & 95.58 \\
  &  SD~2.1    & 98.97 & 97.99 & 98.02 & 89.95 & 93.49 & \textbf{98.26} & 96.11 \\
  &  SD~3.5    & \textbf{99.03} & \textbf{98.30} & 98.25 & \textbf{90.56} & 94.88 & 97.92 & \textbf{96.49} \\
 &   Real  & 99.30 & 97.84 & \textbf{98.60} & 90.50 & \textbf{96.12} & \textbf{98.26} & 96.77 \\
    \bottomrule
\end{tabular}}
\label{tab:vae_ablation}
}
\end{table}

\noindent\textbf{VAE Ablation Study.} We conduct a comprehensive ablation study by evaluating different variants of the VAE on six common STR benchmarks. Specifically, we encode and decode images using each VAE model, and then measure reconstruction quality with PSNR and FID. Furthermore, we assess the impact of VAE reconstruction on downstream text recognition by reporting Acc using a fixed recognizer. For reference, we also report the accuracy using ground-truth (unprocessed) images. 

\begin{table}[ht]
\caption{Ablation study of CFG scale $\omega$ on ScenePair~\cite{zeng2024textctrl}.}
\centering
\begin{tabular}{lccc}
\toprule
\textbf{$\omega$} & \textbf{ACC (\%)$\uparrow$} & \textbf{NED$\uparrow$} & \textbf{FID$\downarrow$} \\
\midrule
1.0    & 59.45  & 0.817 & 60.34 \\
1.5  & 78.98  & 0.934 & \textbf{35.37} \\
1.8  & 82.27  & 0.955 & 37.54 \\
2.0    & 83.83  & 0.961 & 39.69 \\
2.2  & 85.23  & 0.964 & 41.53 \\
2.5  & 86.41  & 0.968 & 43.82 \\
5.0    & \textbf{88.05} & 0.975 & 50.24 \\
7.5  & 87.73  & \textbf{0.976} & 50.72 \\
\bottomrule
\end{tabular}
\label{tab:cfg_ablation}
\end{table}

\noindent\textbf{CFG Scale Ablation Study.}
We conduct an ablation study to investigate the effect of varying the CFG scale $\omega$ during inference. As shown in Tab.~\ref{tab:cfg_ablation}, increasing $\omega$ from 1.0 to 1.5 improves ACC and NED while substantially reducing FID from 60.34 to 35.37. Beyond $\omega=1.5$, further increases generally improve text correctness but worsen FID, revealing a trade-off between accuracy and image fidelity. ACC reaches its highest value at $\omega=5.0$ and decreases slightly at $\omega=7.5$, whereas NED continues to improve. At $\omega=2.2$, the model achieves a balanced performance, with ACC of 85.23\%, NED of 0.964, and FID of 41.53. We therefore select $\omega=2.2$ for subsequent experiments.

\subsection{Additional Editing Evaluation}
\label{app:additional_edit_eval}

\noindent\textbf{Per-image style fidelity beyond FID.}
FID is established as the evaluation metric in MOSTEL~\cite{mostel},
and TextCtrl~\cite{zeng2024textctrl} continues to use the same
evaluation code to report FID on ScenePair, so we keep it for
comparability. For each generated image, we compare it against
its corresponding ScenePair reference image and report LPIPS
(perceptual similarity), SSIM (structural similarity), and PSNR
(pixel-level fidelity). Lower LPIPS and higher SSIM and PSNR
indicate better agreement with the paired reference.
LPIPS measures per-image perceptual and structural similarity,
and FID measures how close the overall image distribution is
to the reference distribution. TextRIRO achieves the best
results on both in Tab.~\ref{tab:per_image_fidelity}, which
indicates that it generates images closer to the real
distribution. PSNR and SSIM, in contrast, are per-image
pixel-level metrics that are sensitive to small spatial shifts,
anti-aliasing, and valid appearance variations in rendered text.
Because the ScenePair reference images are manually collected
and not strictly pixel-aligned pairs, even a more realistic
generation can score lower on these metrics simply because
its pixels do not exactly match the reference.

\begin{table}[ht]
\caption{Per-image fidelity against the ScenePair reference and distributional
fidelity.  Lower LPIPS/FID and higher SSIM/PSNR are better.}
\centering
\begin{tabular}{lcccc}
\toprule
\textbf{Method} & \textbf{LPIPS}$\downarrow$ & \textbf{SSIM}$\uparrow$ & \textbf{PSNR}$\uparrow$ & \textbf{FID}$\downarrow$ \\
\midrule
MOSTEL~\citeyear{mostel} & 0.2947 & 0.3311 & 14.05 & 49.19 \\
TextCtrl~\citeyear{zeng2024textctrl} & 0.3219 & 0.2610 & 11.68 & 43.78 \\
TextSSR~\citeyear{textssr} & 0.2622 & 0.3889 & 14.13 & 45.37 \\
RS-STE~\citeyear{rsste} & 0.2458 & \textbf{0.4347} & \textbf{15.66} & 49.59 \\
TextRIRO & \textbf{0.2394} & 0.3944 & 14.44 & \textbf{41.53} \\
\bottomrule
\end{tabular}
\label{tab:per_image_fidelity}
\end{table}

We acknowledge RS-STE's superiority on these two metrics.
Methodologically, RS-STE builds $L_1$ and LPIPS losses directly
in pixel space, whereas our diffusion-based approach builds
the loss in VAE latent space, which is inherently disadvantaged
at pixel-level reconstruction. Even so, TextRIRO achieves the
second-best results on SSIM and PSNR, demonstrating the
effectiveness of our real-data training paradigm.

\subsection{Additional TextRIRO Details}

\noindent\textbf{Length-agnostic concatenation.}
The even-length regularity originates from the symmetric
concatenation recipe rather than an intrinsic model limitation.
We conduct an additional controlled study on long-text samples
(target length at least 8) to quantify this effect.
We evaluate three models on the long-text set from Multi-Words
(362 odd-length and 401 even-length samples, 763 in total).
The three models differ only in training data. The first is
trained on real STR data (Baseline). The second adds 1M symmetric
data. The third adds 0.5M symmetric and 0.5M asymmetric data
(the asymmetric-augmented recipe).

\begin{table}[ht]
\caption{Odd- vs.\ even-length word accuracy (\%) under
different data recipes.}
\centering
\begin{tabular}{lccc}
\toprule
\textbf{Training data} &
\textbf{Odd length} &
\textbf{Even length} &
\textbf{Overall} \\
\midrule
Real STR data only
& 86.19 & 86.78 & 86.50 \\
+ 1M symmetric (even-only)
& 88.12 & \textbf{90.27} & \textbf{89.25} \\
+ 1M asymmetric-augmented
& \textbf{88.67} & 89.03 & 88.86 \\
\bottomrule
\end{tabular}
\label{tab:length_recipe}
\end{table}

Two observations follow from Tab.~\ref{tab:length_recipe}.
(1) With purely symmetric (even-only) concatenation, the model
still generalizes to odd lengths (+1.93), which confirms that
the even-length regularity is not a hard model limitation.
However, a residual bias remains, as even-length gains (+3.49)
exceed odd-length gains (+1.93).
(2) To eliminate this bias, we introduce an asymmetric
concatenation recipe. We concatenate a text with a variant of
length $L-1$, which supplies odd-length long-text samples while
barely affecting edit-level supervision. Mixing 0.5M asymmetric
with 0.5M symmetric samples (1M total) balances the two gains
(+2.48 odd vs.\ +2.25 even), which removes the even-length
artifact while keeping overall accuracy on par with the
symmetric-only model. The length regularity is a controllable
property of the data recipe.

\noindent\textbf{Complementarity with synthetic sources.}
We also test whether TextRIRO data complements, rather than replaces, existing
synthetic sources.  Tab.~\ref{tab:complementary_data} shows that adding either
TextSSR or ST to TextRIRO and U14M-Filter improves the Union14M-Benchmark
average, and using all sources gives the best result.  TextSSR is particularly
complementary for contextless text, while rendering-based ST helps curved and
multi-oriented cases.  Thus, our real-only construction is a choice to
prioritize realistic editing, not a claim that prior synthetic data is
intrinsically unusable.

\begin{table}[ht]
\centering
\caption{\textbf{Complementarity with synthetic sources.}
Recognition accuracy (\%) of MAERec trained with different data
mixtures.}
\label{tab:complementary_data}
\begingroup
\setlength{\tabcolsep}{2pt}
\renewcommand{\arraystretch}{1.1}
\resizebox{\textwidth}{!}{%
\begin{tabular}{l|c|ccccccc|cccccccc}
\toprule
\multirow{2}{*}{\textbf{Training Data}} &
\multirow{2}{*}{\textbf{Volume}} &
\multicolumn{7}{c|}{\textbf{Common Benchmarks}} &
\multicolumn{8}{c}{\textbf{Union14M-Benchmark}} \\
& &
\textbf{IIIT} &
\textbf{SVT} &
\textbf{IC13} &
\textbf{IC15} &
\textbf{SVTP} &
\textbf{CUTE} &
\textbf{Avg} &
\textbf{CUR} &
\textbf{MLO} &
\textbf{ART} &
\textbf{CTL} &
\textbf{SAL} &
\textbf{MLW} &
\textbf{GEN} &
\textbf{Avg} \\
\midrule
UF (real only)
& 3.22M
& 99.30 & 97.84 & 98.60 & 90.50 & 96.12 & 98.26 & 96.77
& 89.49 & 93.50 & 77.11 & 83.18 & 86.99 & 86.89 & 85.48 & 86.09 \\

TextSSR + UF
& 1M + 3.22M
& 99.07 & 97.53 & \textbf{98.83} & 91.28 & 95.50 & 98.61 & 96.80
& 91.06 & 94.52 & 81.22 & 86.26 & 87.56 & 88.35 & 85.95 & 87.85 \\

TextRIRO + UF
& 1M + 3.22M
& 99.37 & 97.84 & 98.72 & 91.33 & 96.59 & 98.96 & 97.13
& 91.18 & 94.67 & 81.00 & 86.65 & 87.49 & 89.08 & 85.80 & 87.98 \\
\midrule
TextRIRO + TextSSR + UF
& 1M + 1M + 3.22M
& 99.30 & 97.99 & 98.60 & \textbf{91.72} & 96.28 & 98.96 & 97.14
& 91.59 & 94.89 & 81.44 & \textbf{87.93} & 88.19 & 88.83 & 86.22 & 88.44 \\

TextRIRO + ST + UF
& 1M + 1M + 3.22M
& 99.37 & 97.84 & 98.60 & 91.61 & \textbf{97.36} & \textbf{99.65} & 97.40
& \textbf{92.58} & \textbf{95.62} & 81.33 & 86.26 & 87.93
& \textbf{89.81} & 86.22 & 88.54 \\

TextRIRO + TextSSR + ST + UF
& 1M + 1M + 1M + 3.22M
& \textbf{99.40} & \textbf{98.30} & \textbf{98.83}
& 91.44 & 97.21 & 99.31 & \textbf{97.41}
& 92.25 & 95.47 & \textbf{82.00} & 87.14 & \textbf{88.20}
& 89.44 & \textbf{86.32} & \textbf{88.69} \\
\bottomrule
\end{tabular}%
}
\endgroup
\end{table}

\noindent\textbf{Adding paired synthetic data to the editor.}
Our current training uses only real self-paired data.
This framework can equally incorporate existing paired
synthetic data. Under the ablation configuration, we add
the synthetic data provided by MOSTEL~\cite{mostel} for training.

Adding synthetic data raises text ACC and NED, which shows
that synthetic supplementation brings a correctness gain.
However, FID increases, which confirms that synthetic data
has a real distribution gap and hurts realism when added.
To preserve realism (the ``Real Out'' requirement), our paper adopts a real-only training strategy.

\begin{table}[ht]
\caption{Adding paired synthetic data to the self-paired editor
under the ablation configuration.}
\centering
\begin{tabular}{lccc}
\toprule
\textbf{Setting} &
\textbf{ACC (\%)}$\uparrow$ &
\textbf{NED}$\uparrow$ &
\textbf{FID}$\downarrow$ \\
\midrule
Real only (self-paired)
& 82.42 & 0.960 & \textbf{43.59} \\
+ Synthetic data
& \textbf{85.08} & \textbf{0.965} & 49.34 \\
\bottomrule
\end{tabular}
\label{tab:editor_synthetic}
\end{table}

\noindent\textbf{Quality Filtering Bias.} To further validate that the STR model for filtering preserves high-accuracy results and does not compromise dataset quality, we randomly sample 300 instances from the outputs generated by TextRIRO and filtered by SVTRv2~\cite{du2024svtrv2}. We then recruit three assessors, each responsible for evaluating 100 instances. The average accuracy reaches 99.33\% (with individual accuracies of 99\%, 99\%, and 100\%), demonstrating that the retained samples are highly reliable. The filter can nevertheless reject some visually correct samples when the recognizer fails on an unusual style. We choose this conservative trade-off to prioritize label correctness, and report the human check to make this selection bias explicit.

\begin{figure}[ht]
 \centering
 \includegraphics[width=0.8\linewidth]{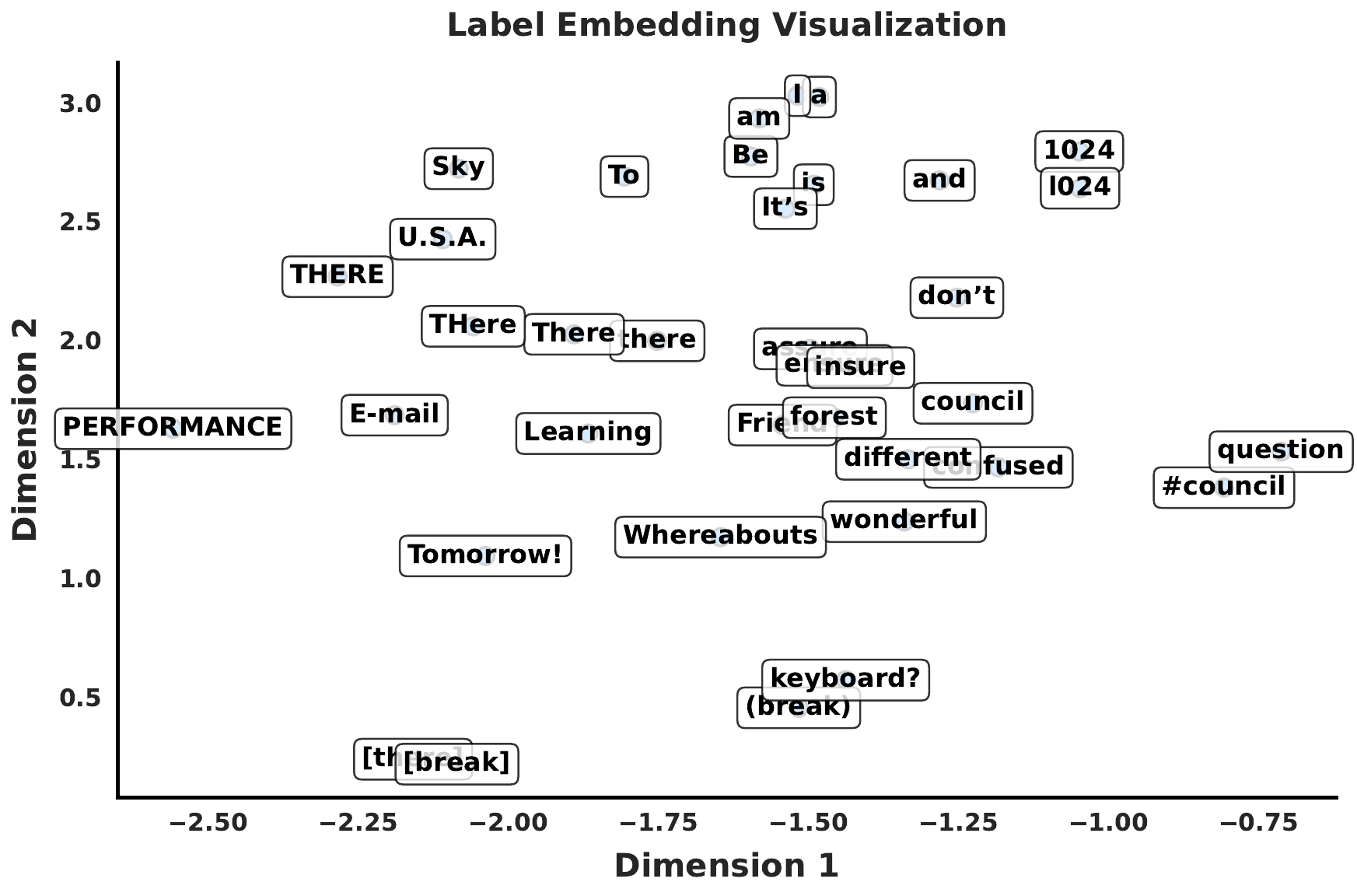}
 \caption{t-SNE~\cite{maaten2008visualizing} visualization of text features by STR pre-trained text encoder.}
 \label{fig:textembd_view}
\end{figure}

\section{Visualization}

\subsection{Character-Aware Embeddings}

The STR pre-trained text encoder endows TextRIRO with strong character awareness. To further evaluate its encoding capabilities, we visualize the embeddings of several text samples using t-SNE~\cite{maaten2008visualizing}, as illustrated in Fig.~\ref{fig:textembd_view}. Overall, words with similar character compositions or comparable text lengths are positioned closer together, while sufficient separation is maintained between distinct samples.  For example, the characters `1' and `l' possess different semantic meanings but exhibit similar glyphs, resulting in `1024' and `l024' being embedded in close proximity. Traditionally, glyph-pre-training is required for such nuanced feature representations, yet STR pre-training achieves this directly.  Additionally, the variations `there', `There', `THere', and `THERE' are embedded near each other, reflecting their shared underlying structure, yet each maintains a distinct position. This separation prevents case confusion in text generation, thereby enabling accurate character-level generation.

\begin{figure}[ht]
 \centering
 \includegraphics[width=0.75\linewidth]{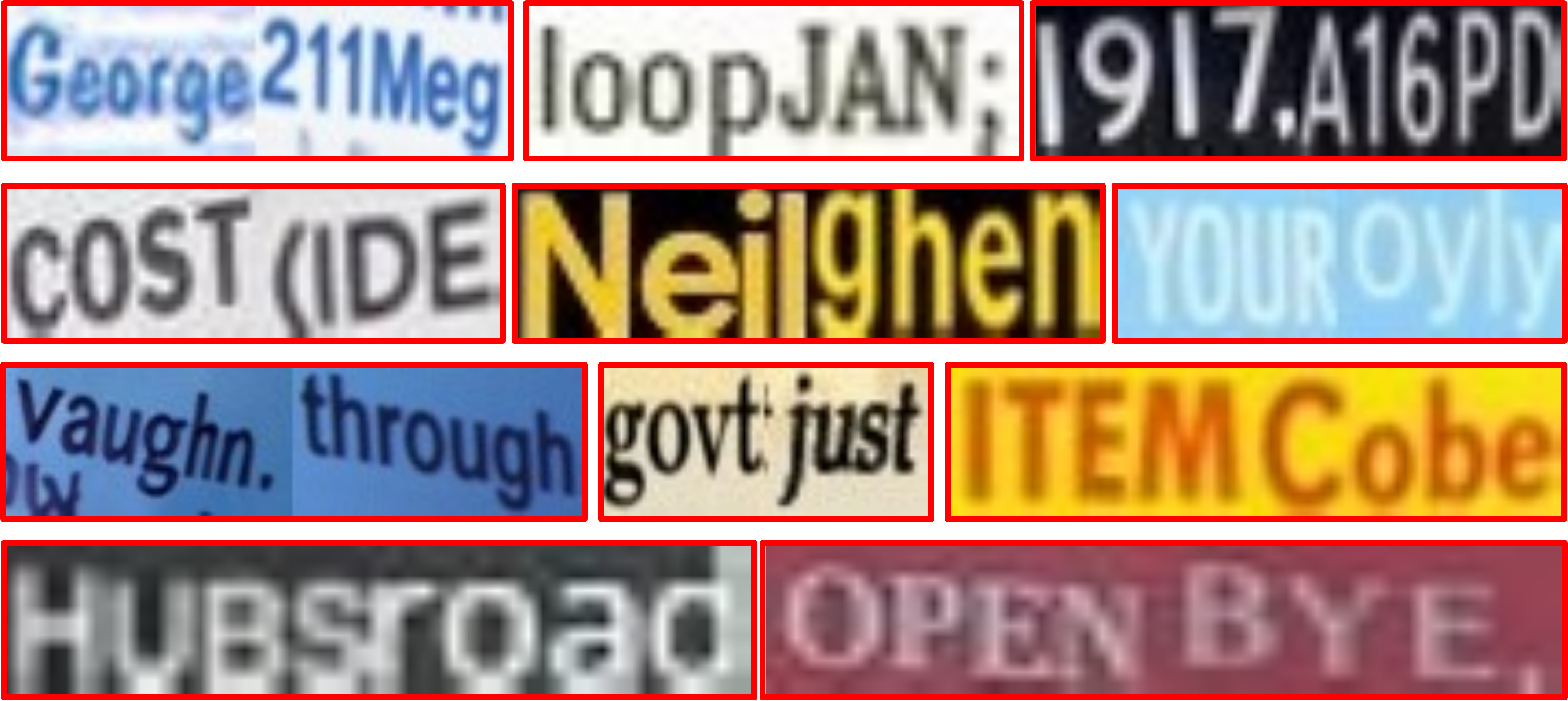}
 \caption{Visualization results randomly sampled from the TextRIRO-3M dataset, which doubles the text length compared to the original scene text images.}
 \label{fig:rirotext_view}
\end{figure}

\begin{figure}[ht]
 \centering
 \includegraphics[width=0.8\linewidth]{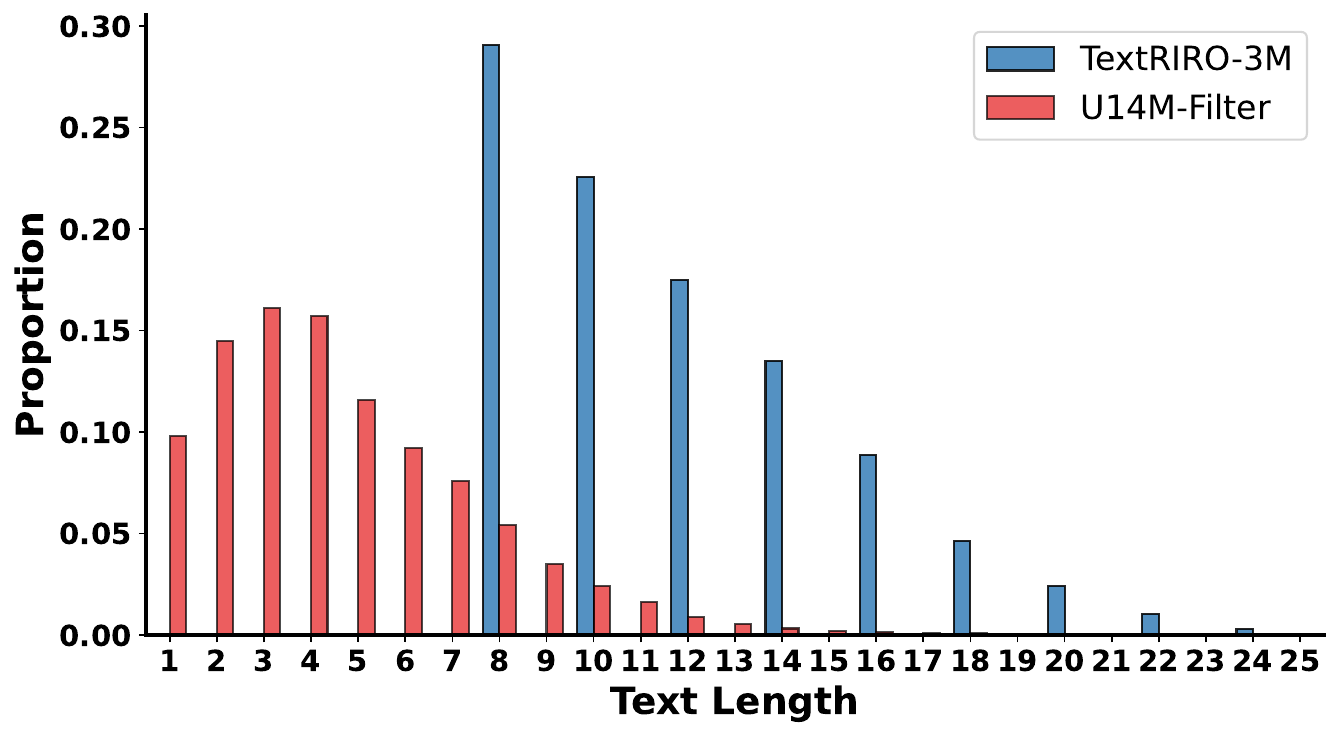}
\caption{Length distributions of real data (U14M-Filter) and the original
symmetric TextRIRO-3M construction.  The symmetric recipe doubles the source
length and hence produces even-length samples, and the asymmetric-augmented recipe
in Tab.~\ref{tab:length_recipe} removes this construction artifact.}
 \label{fig:data_long}
\end{figure}

\subsection{TextRIRO-3M Data View}

Fig.~\ref{fig:rirotext_view} presents several examples from TextRIRO-3M,
demonstrating that the dataset doubles the original text length while
preserving the original style. Fig.~\ref{fig:data_long} further compares the
label length distribution between the original symmetric TextRIRO-3M recipe
and real-world data. The symmetric construction significantly expands coverage
of long text but produces even lengths only. As demonstrated in
Tab.~\ref{tab:length_recipe}, asymmetric augmentation removes this artifact
when broader length coverage is required.

\begin{figure*}[ht]
 \centering
 \includegraphics[width=0.95\linewidth]{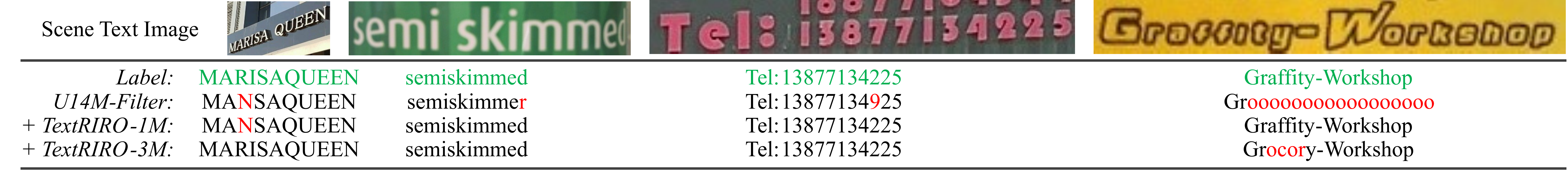}
 \caption{Comparison of STR model (MAERec~\cite{jiang2023revisiting}) prediction outcomes trained on different datasets.}
 \label{fig:rec_result}
\end{figure*}

\subsection{Enhanced STR Visualization}

To clearly illustrate the effectiveness of TextRIRO-3M in improving the performance of STR models under challenging scenarios, especially for long text recognition, we visualize the prediction results of MAERec~\cite{jiang2023revisiting} trained with different datasets in Fig.~\ref{fig:rec_result}. The experimental results show that, after incorporating TextRIRO-3M during training, the model produces more accurate and robust predictions. In particular, it demonstrates improved recognition of long and complex text sequences, effectively reducing error rates in difficult cases.

\subsection{Bad Cases and Discussion}

\begin{figure}[ht]
 \centering
 \includegraphics[width=0.8\linewidth]{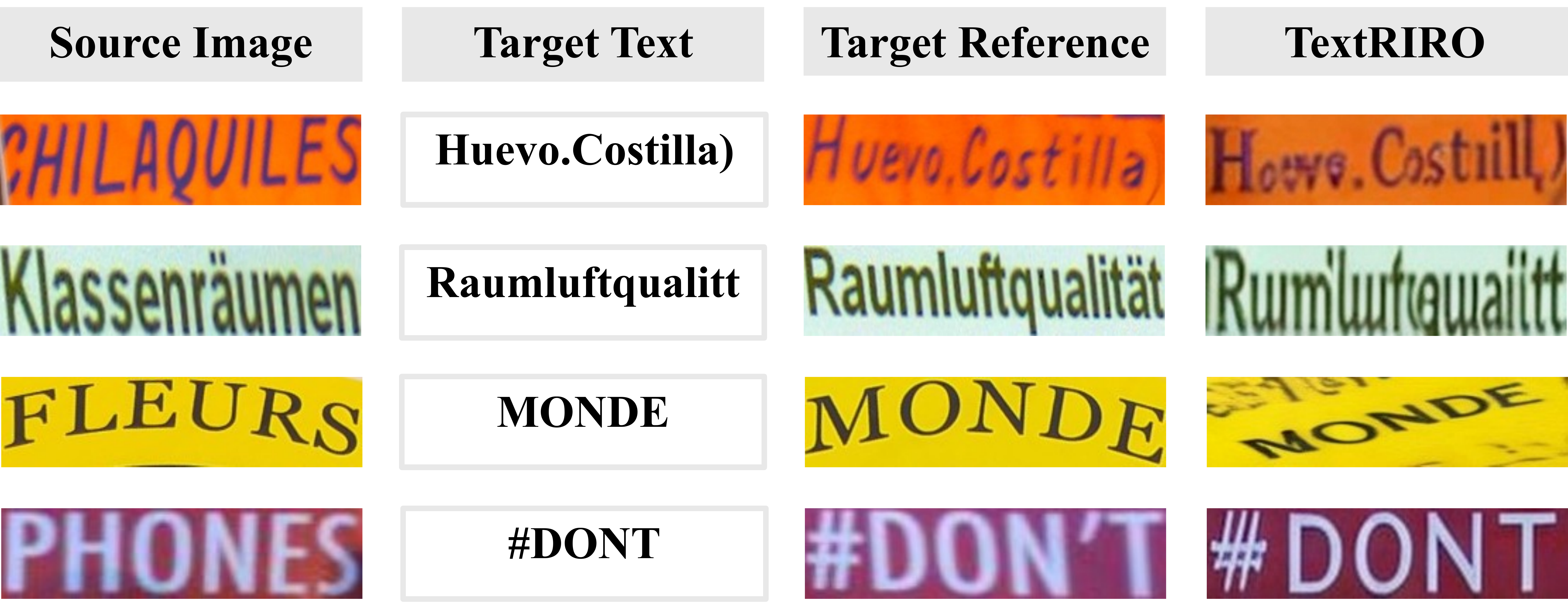}
 \caption{Bad Cases. We show several disappointing editing results produced by TextRIRO.
}
 \label{fig:badcase}
\end{figure}

It is important to acknowledge that our STE approach is not without limitations. Fig.~\ref{fig:badcase} illustrates three common failure cases, along with the corresponding causes and potential solutions:
\begin{enumerate}
    \item \textbf{Long Text:} The text encoder used in our training pipeline is pre-trained from the STR model, which is primarily trained on datasets containing short text, resulting in limited capacity for capturing long text dependencies. Additionally, TextRIRO is trained on the same datasets, further constraining its ability to handle long text. To address this, we are inspired by the construction of the TextRIRO-3M dataset: we segment long text into shorter fragments, synthesize each fragment individually, and then concatenate these fragments to form the final long text. This approach leverages the inherent strengths of short text synthesis while enabling robust long text generation.
    
    \item \textbf{Curved Text and Complex Geometry:} As shown in Fig.~\ref{fig:badcase}, TextRIRO recognizes that the input is not standard horizontally aligned text, but may interpret highly curved text as slanted. Crop preserves local tilt and curvature, and near-vertical inputs are rotated in preprocessing. Consequently, TextRIRO achieves equal-/variable-length accuracy of 64.87/48.12 on CUTE and 79.69/42.31 on SVTP, outperforming the compared baselines. Tab.~\ref{tab:layout_attribution} further isolates the operator contribution. However, extreme curvature and large geometric deformation remain difficult because regular layouts dominate the training data. Future work should incorporate finer-grained control points, similar to TPS modules in STR, to enable more precise geometric control.
    
    \item \textbf{Special Symbols and Languages:} The synthesis of special symbols remains challenging because of both sparse training coverage and the current vocabulary/data coverage. This also restricts cross-lingual generalization: the self-pairing operation itself is language-agnostic, but sufficiently large real labeled datasets are currently much more available for Latin scripts than for low-resource languages. Collecting balanced real labeled data with broader character and language coverage is necessary for stable multilingual generation. TextRIRO will also provide more abundant synthetic data for multilingual and general OCR~\cite{ye2026all, du2026unirec}.

\end{enumerate}

\noindent\textbf{Style preservation and disentanglement.}
Shuffle preserves stroke thickness because it permutes equal-width columns
without rescaling them.  Crop can change apparent stroke scale, but the two
operators are complementary: pure-Shuffle instances remain in training with
non-trivial probability (approximately 20\% for an average label length of
five) and anchor stroke-width preservation, while Crop supplies the scale
variation needed for variable-length editing.  The style encoder is not given
an explicit transformation-type label, and it learns from the mixed perturbation
distribution.  Together with the disrupted semantics in the source and the
separate text condition, this prevents simple content copying, as further
supported by the None ablation described above.  These mechanisms do not
eliminate every failure case, but explain why the combined operator is more
robust than either component alone.

\section{Responsible Research Notes}
\label{app:responsible}

\noindent\textbf{Broader impacts and safeguards.} TextRIRO can support benign applications such as visual content localization, accessibility, and realistic STR data generation, but scene text editing can also be misused to falsify text in images. We therefore intend to release code, data-construction scripts, and TextRIRO-3M with clear research-use and acceptable-use notices, retain STR-based quality filtering, and encourage provenance tracking or watermarking when the technology is deployed.

\noindent\textbf{Existing assets and licenses.} The work uses publicly available academic datasets, model components, and baselines, all cited in the main paper and bibliography. Users of released artifacts will be instructed to follow the original licenses and terms for Union14M, U14M-Filter, ScenePair, the STR benchmarks, Stable Diffusion components, and compared open-source STE models.

\noindent\textbf{New assets.} TextRIRO-3M is a newly constructed derived dataset. The paper documents its source data, generation pipeline, STR filtering, scale, and length distribution in Sec.~\ref{sec_3-4}, Sec.~\ref{sec:ablation}, Tab.~\ref{tab:comp_data}, Tab.~\ref{tab:comp_data_model}, and Fig.~\ref{fig:data_long}.

\noindent\textbf{Human assessment protocol.} The quality-filtering bias check in the Appendix used three assessors to judge generated text-image pairs. The instruction was: ``Given a generated image and its target label, mark the sample correct only if all visible characters match the target label in order.'' No personal data were collected, no crowdsourcing platform was used, no external paid workers were recruited, and the task involved only visual inspection of scene-text images.

\end{document}